\documentclass[12pt]{article}
\usepackage{inputenc}
\usepackage{amssymb}
\usepackage{graphicx}
\usepackage{caption}
\usepackage{subcaption}
\usepackage{amsmath}
\usepackage{url}
\usepackage{float}
\usepackage{epsfig}
\usepackage{setspace}

\newcommand{\be}{\begin{equation}}
\newcommand{\ee}{\end{equation}}

\begin{document}
\date{}
\title{Learning in the Machine: the Symmetries of the Deep Learning Channel}
\author{ Pierre Baldi$^{1,}${\footnote{Corresponding author.$^1$ Department of Computer Science, University of California, Irvine.
 $^2$ Department of Mathematics, University of California, Irvine.}, Peter Sadowski$^1$}, and Zhiqin Lu$^2$}

\maketitle

\begin{abstract}
{\bf Abstract:}  
In a physical neural system, learning rules must be local both in space and time. In order for learning to occur, non-local information must be communicated to the deep synapses through a communication channel, the deep learning channel.
We identify several possible architectures for this learning channel (Bidirectional, Conjoined, Twin, Distinct) 
and six symmetry challenges: 1) symmetry of architectures;
2) symmetry of weights; 3) symmetry of neurons; 4) symmetry of derivatives; 5) symmetry of processing; and 6) symmetry of learning rules. Random backpropagation (RBP) addresses the second and third symmetry, and some of its variations, such as skipped RBP (SRBP) address the first and the fourth symmetry. Here we address the last two desirable symmetries showing  through simulations that they can be achieved and that the learning channel is particularly robust to symmetry variations. Specifically, random backpropagation and its variations can be performed with the same non-linear neurons used in the main input-output forward channel, and the connections in the learning channel can be adapted using the same algorithm used in the forward channel, removing the need for any specialized hardware in the learning channel. Finally, we provide  mathematical results in simple cases showing that the learning equations in the forward and backward channels converge to fixed points, for almost any initial conditions.
In symmetric architectures, if the weights in both channels are small at initialization, adaptation in both channels leads to weights that are essentially symmetric during and after learning. Biological connections are discussed.
\end{abstract}

\section{Introduction}

Backpropagation implemented in digital computers has been successful at addressing a host of difficult problems 
ranging from computer vision 
\cite{krizhevsky2012imagenet,szegedy2015going,srivastava2015training,he_delving_2015} to speech recognition \cite{graves2013speech} in engineering, and from high energy physics \cite{baldi_searching_2014,baldidarkmatter15}
to biology \cite{deepcontact2012,zhou2015predicting,baldiagostinelli2016} in the natural sciences. Furthermore, recent results have shown that 
backpropagation is optimal in some sense \cite{baldi2016local}.
However, backpropagation implemented in digital computers is not the real thing. It is merely a digital emulation of a learning process occurring in an idealized 
physical neural system. Thus thinking about learning in this digital simulation can be useful but also misleading, as it often obfuscates fundamental issues. Thinking about learning in physical neural systems or learning in the machine--biological or other--is useful not only for better understanding how specific or idealized machines can learn, but also to better understand fundamental, hardware-independent, principles of learning. And, in the process, it may occasionally also be useful for deriving new approaches and algorithms to improve the effectiveness of digital simulations and current applications.

Thinking about learning in physical systems first leads to the notion of locality \cite{baldi2016local}. In a physical system, a learning rule for adjusting synaptic weights can only depend on variables that are available locally in space and time. This in turn immediately identifies a fundamental problem for backpropagation in a physical neural system and leads to the notion of a learning channel. The critical equations associated with backpropagation show  that the deep weights of an architecture must depend on non-local information, such as the targets. Thus a channel must exist for communicating this information to the deep synapses--this is the learning channel \cite{baldi2016local}.

Depending on the hardware embodiment, several options are possible for implementing the learning channel. A first possibility is to use the forward connections in the reverse direction. A second possibility is to use two separate channels with different characteristics and possibly different hardware substrates in the forward and backward directions. These two cases will not be further discussed here. The third case we wish to address here is when the learning channel is a separate channel but it is similar to the forward channel, in the sense that it uses the same kinds of neurons, connections, and learning rules. Such a learning channel is faced with at least six different symmetry challenges:
1) symmetry of architectures;
2) symmetry of weights; 3) symmetry of neurons; 4) symmetry of derivatives; 5) symmetry of processing; and 6) symmetry of learning rules, where in each case the corresponding symmetry is in general either desirable (5-6) or undesirable (1-4).
 
In the next sections, we first identify the six symmetry problems and then show how they can be addressed within the formalism of simple neural networks. While biological neural networks remain the major source of  inspiration for this work, the analyses derived are more general and not tied to neural computing in any particular substrate.

\section{The Learning Channel and the Symmetry Problems}

\subsection{Basic Notation}
Throughout this paper, we consider layered feedforward neural network architectures and supervised learning tasks. We will denote such an architecture by

\be
{\cal A}[N_0,\ldots, N_h, \ldots ,N_L]
\label{eq:arc}
\ee
 where $N_0$ is the size of the input layer, $N_h$ is the size of hidden layer $h$, and $N_L$ is the size of the output layer. For simplicity, we assume that the layers are fully connected and let $w^h_{ij}$ denote the weight connecting neuron $j$ in layer $h-1$ to neuron $i$ in layer $h$. The output
$ O_i^h $ of neuron $i$ in layer $h$ is computed by:

\be
O_i^h=f_i^h(S_i^h) \quad {\rm where} \quad S_i^h=\sum_j w_{ij}^h O^{h-1}_j
\label{eq:neuron}
\ee
The transfer functions $f_i^h$ are usually the same for most neurons, with typical exceptions for the output layer, and usually are monotonic increasing functions.
Typical functions used in artificial neural networks are: the identity, the logistic function, the hyperbolic tangent function, the rectified linear function, and the softmax function.

We assume that there is a training set of $M$ examples consisting of input-target pairs $(I(t), T(t))$, with $t=1,\ldots, M$. $I_i(t)$ refers to the $i$-th component of the $t$-th training example, and similarly for $T_i(t)$. In addition there is an error function $\cal E$ to be minimized by the learning process. In general, we will assume standard error functions, such as the squared error in the case of regression problems with identity transfer functions in the output layer, or relative entropy in the case of classification problems  with logistic (two-class) or softmax (multi-class) transfer functions in the output layer, although this is not an essential point. The error function is a differentiable function of the weights and its critical points are given by the equations
$\partial {\cal E}/\partial w^h_{ij}=0$.

\subsection{Local Learning}

In a physical neural system, learning rules must be local \cite{baldi2016local}, in the sense that they can only involve variables that are available locally in both space and time, although for simplicity here we will focus primarily on locality in space. 
Thus typically, in the present formalism, a local learning rule for a deep layer is of the form:

\be
\Delta w_{ij}^h=F(O_i^h,O_j^{h-1},w_{ij}^h)
\label{eq:local1}
\ee
while for the top layer:

\be
\Delta w_{ij}^L=F(T_i,O_i^L,O_j^{L-1},w_{ij}^L)
\label{eq:local2}
\ee
assuming that the targets are local variables for the top layer. Hebbian learning \cite{hebb1949organization} is a form of local learning. Deep local learning corresponds to stacking local learning rules in a feedforward neural network. 
Deep local learning using Hebbian learning rules has been proposed by Fukushima \cite{fukushima1980neocognitron} to train the neocognitron architecture, essentially a feed forward convolutional neural network inspired by 
the earlier neurophysiological work of Hubel and Wiesel \cite{hubel1962receptive}. However, in deep local learning, information about the targets cannot be propagated to the deep layers and therefore in general deep local learning cannot
find solutions of the critical equations, and thus cannot succeed at learning complex functions in any optimal way.

\subsection{The Learning Channel}

Ultimately, for optimal learning, all the information required to reach a critical point of $\cal E$ must appear in the learning rule of the deep weights.
Setting the gradient (or the backpropagation equations) to zero shows immediately that in general at a critical point all the deep synapses must depend on the target or the error information, and this information is not available locally 
\cite{baldi2016local}. Thus, to enable efficient learning, there must exist a communication channel to communicate information about the targets or the errors to the deep weights. This is the deep learning channel or, in short, the learning channel. Note that the learning channel is different from the typical notion of ``feedback.'' Although feedback and learning may share the same physical connections, these refer in general to two different processes  that often operate at very different time scales, the feedback being fast compared to learning. 

In a learning machine, one must think about the physical nature of the channel.
A first possibility is to use the forward connections in the reverse direction. This is unlikely to be the case in biological neural systems, in spite of known example of retrograde transmission, as discussed later in Section 6. 
A second possibility is to use two separate channels with different characteristics and possibly different hardware substrates in the forward and backward directions. As a thought experiment, for instance, one could imagine using electrons in one direction, and photons in the other. Biology can easily produce many different types of cells, in particular of neurons, and conceivably it could use special kinds of neurons in the learning channel, different from all the other neurons. 
While this scenario is discussed in Section 6, in general it does not seem to be the most elegant or economical solution as it requires different kinds of hardware in each channel.
In any case, regardless of biological considerations, we are interested here in exploring the case where the learning channel is as similar as possible to the forward channel, in the sense of being made of the same hardware, and not requiring any special accommodations. However, at the same time, we also want to get rid of any undesirable symmetry properties and constraints, as discussed below. This leads to six different symmetry challenges, four undesirable and two desirable ones. 

\subsection{The Symmetry Problems}

\par\null\par
\noindent{\bf Symmetry of Architectures [ARC]:}
Symmetry of architectures refers to having the exact same architecture in the forward and in the backward channel, with the same number of neurons in each hidden layer and the same connectivity. This corresponds to the Bidirectional, Conjoined, and Twin cases defined below. In the Bidirectional and Conjoined case the Symmetry of Architectures is even stronger, in the sense that the same neurons are used in the forward and the backward channel.
ARC is very constraining in a physical system, and it would be desirable if this constraint was unnecessary. 

\par\null\par
\noindent{\bf Symmetry of Weights (Transposition)[WTS]:}
This is probably the most well known symmetry. In the backpropagation equations, the weights in the learning channel are identical transposed copies  of the weights in the forward network. This is a special and even stronger case of architectural symmetry. Furthermore, such a constraint would have to be satisfied not only at the beginning of learning, but it would have to be maintained also at all times throughout any learning process. This poses a major challenge in any physical implementation, including biological ones, and may thus be considered undesirable. If symmetry of the weights is required, then a physical mechanism must be proposed by which such symmetry could be achieved. As we shall see, approximate symmetry can arise automatically under certain conditions.

\par\null\par
\noindent{\bf Symmetry of Neurons (Correspondence)[NEU]:}
For any neuron $i$ in layer $h$, backpropagation computes a backpropagated error $B_i^h$. If $B_i^h$ is computed in a separate learning channel, how does the learning channel know that this variable correspond to neuron $i$ in layer $h$ of the forward pathway? Thus there is a correspondence problem between variables computed in the learning channel and neurons in the forward channel. A desirable solution would have to address this question
in a way that does not violate the locality principle and other constraints of a learning machine.

\par\null\par
\noindent{\bf Symmetry of Derivatives (Derivatives Transport and Correspondence) [DER]:}
Each time a layer is traversed, each backpropagated error must be multiplied by the derivative of the activation of the corresponding forward neuron. Again, how does the learning channel, as a separate channel, know about all these derivatives, and which derivatives correspond to which neurons? A desirable solution would have to address this question in way that does not violate the locality principle and other constraints of a learning machine.

\par\null\par
\noindent{\bf Symmetry of Processing (Non-Linear vs Linear) [LIN]:}
The backpropagation equations are linear in the sense that they involve only multiplications and additions, but no non-linear transformations. Thus a straightforward implementation of backpropagation would require non-linear neurons in the forward channel and linear neurons in the learning channel. Having different kinds of neurons, or neurons that can operate in different regimes is possible, but not particularly elegant, and it would be desirable to be able to use the same neurons in both channels. Since non-linear neurons are necessary in the forward channel to implement non-linear input-output functions, the question we address here is whether we can have similar non-linear neurons in the learning channel.

\par\null\par
\noindent{\bf Symmetry of Adaptations and Learning Rules [ADA]:}
Finally, in backpropagation, a neuron in the forward networks adapts its incoming weights using the learning rule: $\Delta w_{ij}^h=\eta B_iO_j^{h-1}$ where $O_j^{h-1}$ is the activity of the presynaptic neuron, and $B^h_i$ is the postsynaptic backpropagated error. All the weights in the forward network evolve in time during learning. If the learning channel is made of the same kinds of neurons, shouldn't the weights in the learning channel adapt too, and preferably using a similar rule? This is also desirable otherwise one must postulate the existence of at least two types of neurons or connections, those that adapt and those that do not, and use each type exclusively in the forward and in the backward channel respectively.

\par\null\par
\noindent{\bf Other Symmetries:}
While the symmetries above are the major symmetries to be considered here, in a physical system there exist other properties that can be investigated for symmetry or similarity between the forward and the learning channel. Some of these will be considered too, but more briefly. For instance, are there similar kinds of noise and noise levels in both channels?
Can dropout  \cite{srivastava_dropout_2014,baldidropout14} be used in both channels?
Is the precision on the weights the same in both channels? Another asymmetry between the channels, left for future work, is that the forward channel has a target whereas the learning channel does not.
Finally, it must be noted that in backpropagation neurons operate in fundamentally different ways in the forward and backward directions. In particular, in backpropagation the backpropagated error is never added to the input activation in order to trigger a neuronal response.
Thus the standard backpropagation model assumes that neurons can distinguish the forward messages from the backward messages and react differently to each. While one can imagine plausible mechanisms for doing that, it may also be desirable to come up with models where the two kinds of messages are treated in the same way, and the backpropagated message is included in the total neuronal activation. A small step in this direction is taken in Section 3.6.

Solutions for the first four symmetry problems are provided to some extent by the study of random backpropagation and several of its variations \cite{lillicrap2016random,baldiRBP2016AI}, which we now briefly describe.

\section{Backpropagation, Random Backpropagation, and their Variants}

\subsection{Backpropagation (BP)}

Standard backpropagation implements gradient descent on $\cal E$, and can be applied in a stochastic fashion on-line (or in mini batches) or in batch form, by summing or averaging over all training examples.
For a single example, omitting the $t$ index for simplicity, the standard backpropagation learning rule is given by:

\be
\Delta w_{ij}^h=-\eta \frac{ \partial {\cal E}}{\partial w_{ij}^h}=\eta B_i^hO_j^{h-1}
\label{eq:bp}
\ee
where $\eta$ is the learning rate, $O_j^{h-1}$ is the presynaptic activity, and 
$B_i^h$ is the backpropagated error. Using the chain rule, it is easy to see that the backpropagated error satisfies the recurrence relation:

\be
B_i^h=\frac { \partial {\cal E}}{\partial S^h_i}=(f_i^h)' \sum_k  w^{h+1}_{ki} B^{h+1}_k
\label{eq:bp}
\ee
with the boundary condition:

\be
B_i^L=\frac{\partial {\cal E}_i}{\partial S^L_i}=T_i-O^L_i
\label{eq:bp1}
\ee
Thus in backpropagation the errors are propagated backwards in an essentially linear fashion, using the transpose of the forward matrices, hence the symmetry of the weights, with a multiplication by the derivative of the corresponding forward activations every time a layer is traversed.

\subsection{Random Backpropagation (RBP)}

Standard random backpropagation \cite{lillicrap2016random}
operates exactly like backpropagation except that the weights used in the backward pass are completely random and fixed. Thus the learning rule becomes:

\be
\Delta w_{ij}^h=\eta R_i^h O_j^{h-1}
\label{eq:bp}
\ee
where the randomly backpropagated error satisfies the recurrence relation:

\be
R_i^h= (f_i^h)' \sum_k  c^{h+1}_{ik} R^{h+1}_k
\label{eq:bp}
\ee
where the weights $c^{h+1}_{ik}$ are random and fixed.
The boundary condition at the top remains the same:

\be
R_i^L=\frac{\partial {\cal E}_i}{\partial S^L_i}=T_i-O^L_i
\label{eq:bp1}
\ee
Note that as described, RBP solves the second symmetry problem, but not the other five symmetry problems.

\subsection{Skipped Random Backpropagation (SRBP)}

Skipped random backpropagation was introduced independently in
\cite{NIPS2016_6441,baldiRBP2016AI}. In its basic form, SRBP uses connections with random weights 
that run directly from the top layer to each deep neuron. In this case, the signal carried by the learning channel has the form:

\be
R_i^h= (f_i^h)' \sum_k  c^{h}_{ik} (T-O^L)_k
\label{eq:bp}
\ee
where $c^h_{ik}$ are fixed random weights. SRBP has been shown, both through simulations and mathematical analyses, to work well even in very deep networks. Furthermore, another important conclusion derived from the study of SRBP, is that
when updating the weight $w_{ij}^h$, the only derivative information that matters is the derivative of the activation of neuron $i$ in layer $h$, and this information is available locally. Information about all the other derivatives, which is carried by the backpropagated signal $B_i^h$ in standard backpropagation, is not local and is not necessary for successful learning. Note that omitting all the derivatives does not work
\cite{baldiRBP2016AI}.

\subsection{Other Variants: Adaptation}

Several other variants are considered in 
\cite{baldiRBP2016AI}. The most important one for our purposes is the adaptive variants of RBP and SRBP, called ARBP and ASRBP. In these variants, the random weights of the learning channel are adapted 
using the product of the corresponding forward and backward signals, so that $\Delta c_{rs}^l = \eta R_s^{l+1} O_r^{l} $, where $R$ denotes the randomly backpropagated error. While ARBP and ASRBP 
allow both channels to learn, and use rules that are roughly similar, these rules are not identical.
This is because in ARBP and ASRBP, and all the previously described algorithms, propagation in the learning channel is linear, as opposed to the non-linear propagation in the forward channel. As a result, derivatives of activations appear in the learning rules for the forward weights, but not for the weights in the learning channel. In this work we also explore the case where the learning channel is non-linear too and modify its learning rule accordingly by including the derivatives of the activations in the learning channel.

\begin{figure}[h!]
    \centering
    \includegraphics[width=1.0\textwidth]{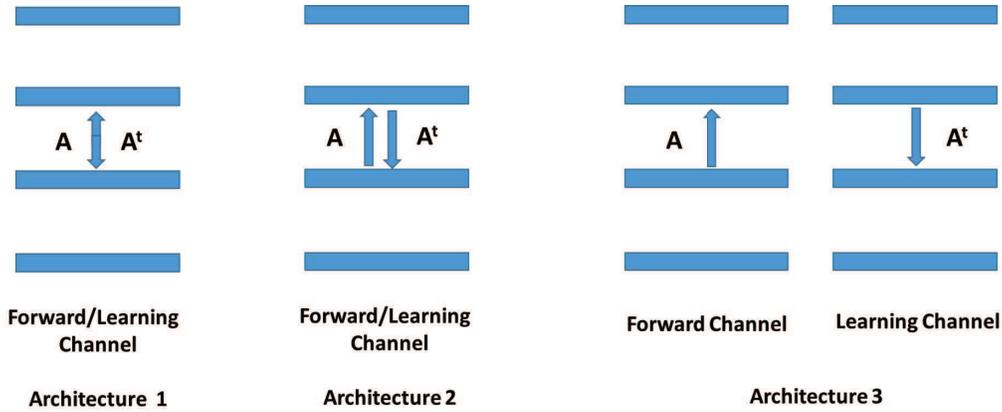}
    \caption{Representations of three different physical implementations of standard backpropagation (BP). Architecture 1 correspond to the Bidirectional Case, where information can flow in both directions along the same connections.
Architecture 2 corresponds to the Conjoined Case, where the architecture and the neurons in the learning channel are identical to those in the forward channel. Architecture 3 corresponds to the Twin Case where the architecture of the learning channel is identical to the forward channel but the neurons are different. 
    All the symmetry problems are evident in Architecture 3: How can the architecture of the learning channel be identical to the forward architecture? How can the weights in the learning channel be exactly symmetric to the weights in the forward channel? How can a backpropagated error computed in the learning channel precisely reach the corresponding neuron in the forward channel? How can the learning channel know about the derivatives of the activation functions in the forward channel? Why is the forward channel processing data in a non-linear fashion while the learning channel processes data in a linear fashion? And why is the forward channel adaptive but the learning channel is not? 
}
    \label{fig:BP-ARCH123}
\end{figure}

\begin{figure}[h!]
    \centering
    \includegraphics[width=0.60\textwidth]{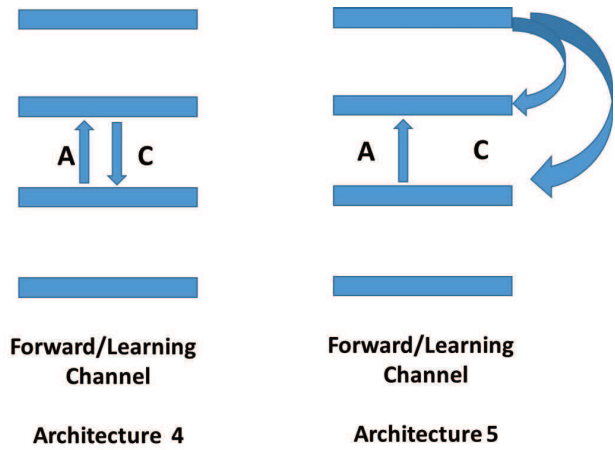}
    \caption{Representations of physical implementations of standard random backpropagation (RBP) and skipped random backpropagation (SRBP) in the Conjoined Case. Standard RBP and SRBP addresse the WTS problem. The ARC, NEU, and DER problems are addressed automatically by the conjoined nature of the architecture.     
}
    \label{fig:RBP-ARCH45}
\end{figure}

\begin{figure}[h!]
    \centering
    \includegraphics[width=0.95\textwidth]{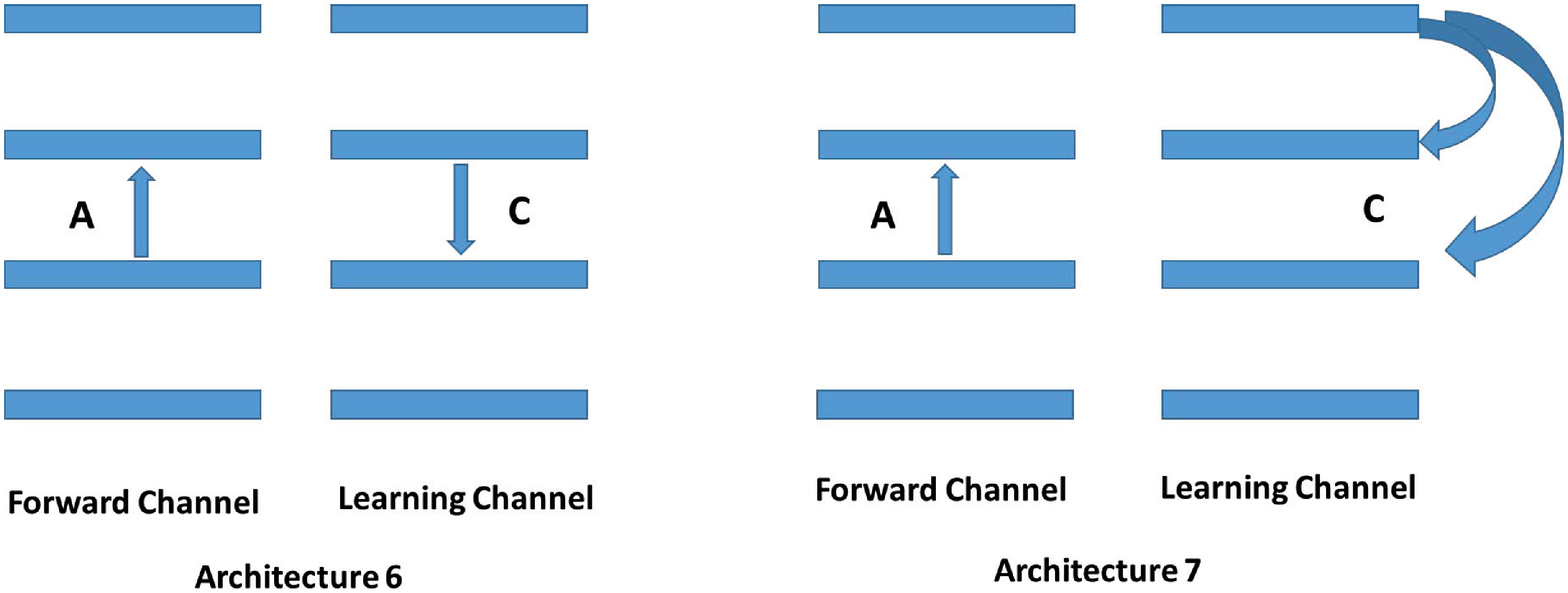}
    \caption{Representations of physical implementations of standard random backpropagation (RBP) and skipped random backpropagation (SRBP) using a learning channel which is distinct from the forward channel (Twin Case). 
    }
    \label{fig:RBP-ARCH67}
\end{figure}

\subsection{Addressing the Symmetry Problems}

\begin{figure}[h!]
    \centering
    \includegraphics[width=0.95\textwidth]{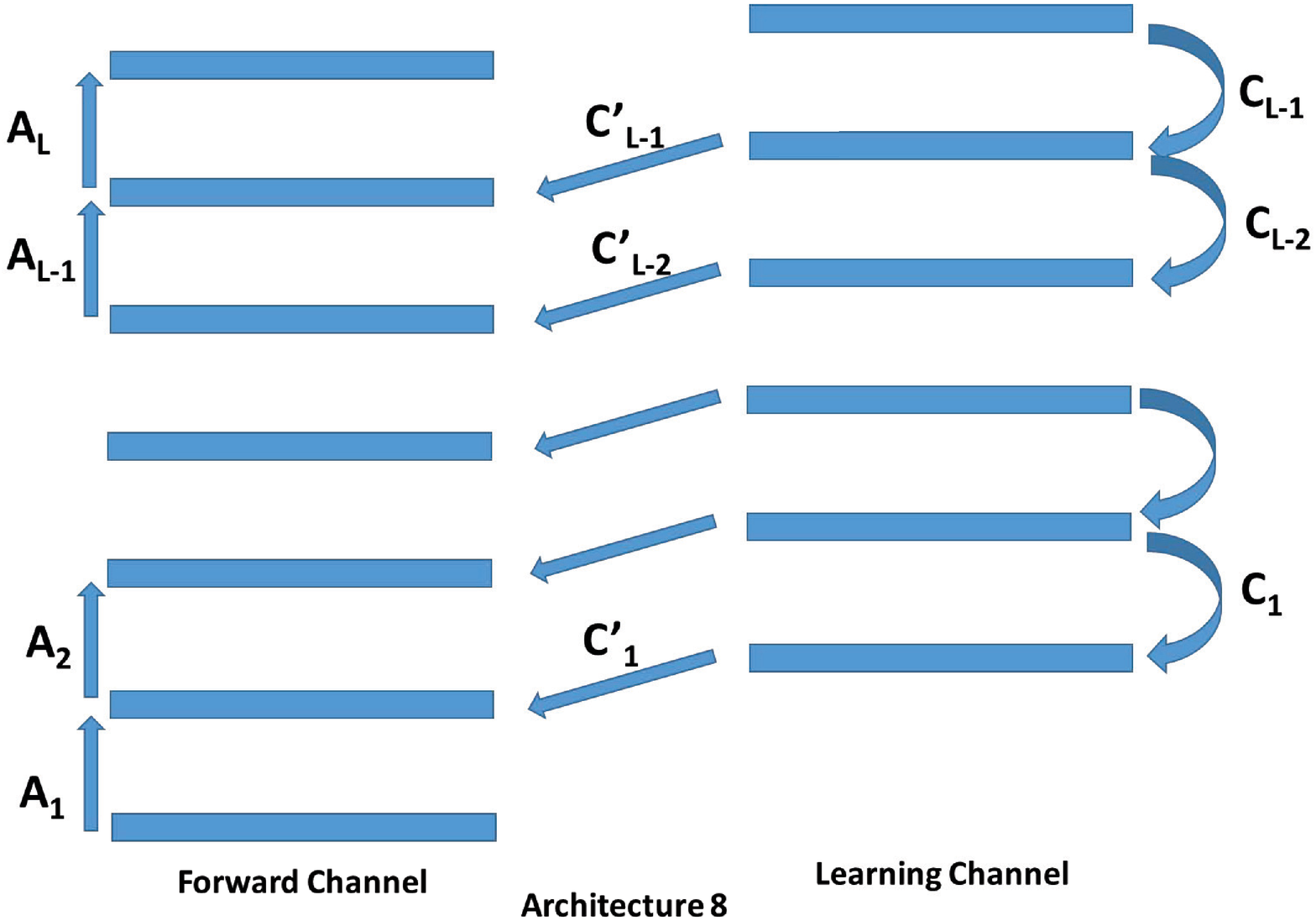}
    \caption{Random backpropagation with random matrices ($C'$) also connecting the learning channel to the corresponding layer in the forward channel (Non-Identical Twin Case, or Distinct Case). This version addresses both the symmetry of the weights problem, and the neuronal correspondence problem. In addition, insights from SRBP show that only local information about the derivative of the activation function of the neuron under consideration for learning is needed(i.e. the derivatives in the upper layers are not needed). So this version can also address the issue of the transport of the derivatives from one channel to the other--such transport is not necessary. In this version, the learning channel can be run in linear or non linear fashion. The only two remaining symmetry problems that this version does not address are the symmetry in architectures and the symmetry in adaptability of the two channels.
    }
    \label{fig:RBP-ARCH8}
\end{figure}

\begin{figure}[h!]
    \centering
    \includegraphics[width=0.95\textwidth]{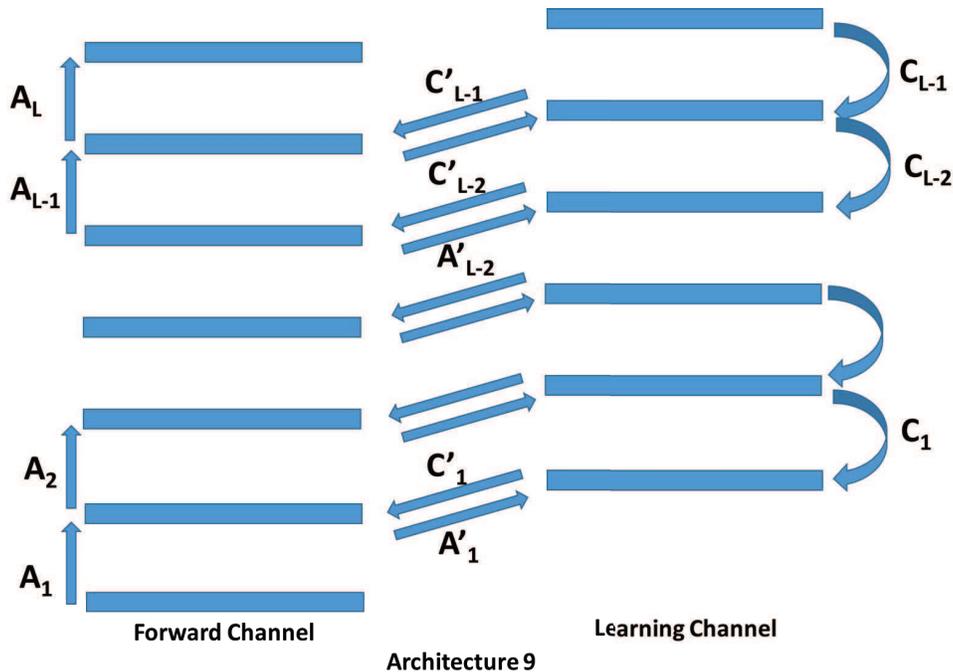}
    \caption{Random backpropagation with random matrices ($C'$) also connecting the learning channel to the corresponding layer in the forward channel, and random matrices ($A'$) connecting the forward channel to the corresponding layer in the learning channel (Non-Identical Twin Case and Distinct Case). This version solves the correspondence problem in the reverse direction, allowing the forward channel to provide ``targets'' for the learning channel. Thus the learning channel can adapt by using the exact same learning rule as the forward channel. The only symmetry problem that is not addressed is the symmetry in architectures.
    }
    \label{fig:RBP-ARCH9}
\end{figure}

\begin{figure}[h!]
    \centering
    \includegraphics[width=0.95\textwidth]{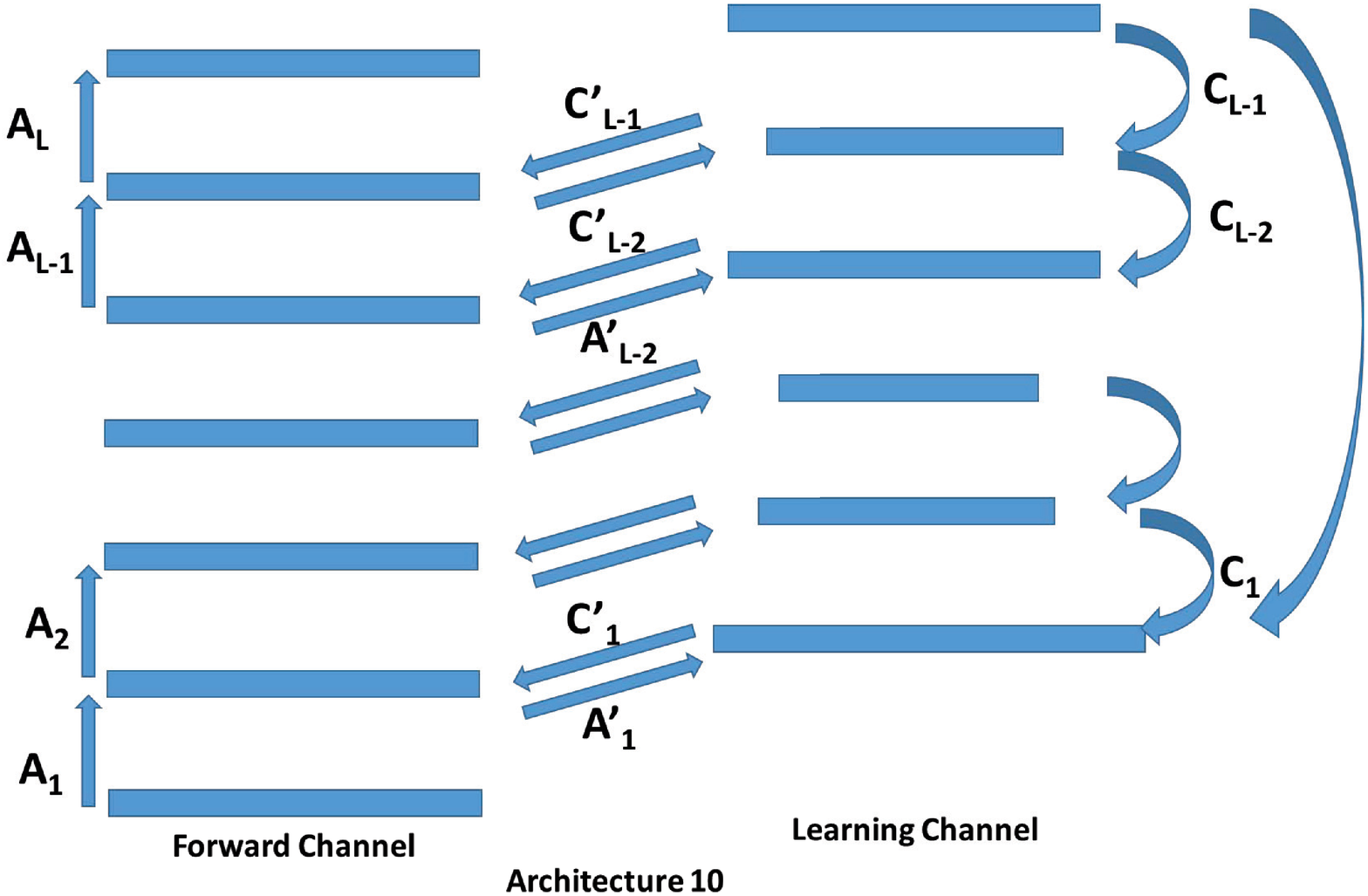}
    \caption{This configuration addresses all six symmetry problems (Distinct Case). Not only the learning channel can have a different architecture but it is also allowed to have skip connections of various kinds.
    }
    \label{fig:FinalRBP}
\end{figure}

We now review how these algorithms address some of the symmetry problems, but only partially, in relation to the corresponding architectures described in Figures \ref{fig:BP-ARCH123}-\ref{fig:FinalRBP}.
In these figures, the symbol $A$ represents the forward matrices, $A^t$ the transpose of the forward matrices, and $C$ the random matrices.

\begin{itemize}
\item {\bf Architecture 1:} This represents a physical implementation of BP where information can flow bidirectionally along the same connections
(Figure \ref{fig:BP-ARCH123}).
In this Bidirectional Case, the ARC, WTS, NEU, and ADA problems are solved by definition (assuming the ``weight'' on a connection is the same in both directions) and so is the DER problem. However the LIN problem is not addressed--neurons would have to operate differently in the two directions--and bidirectional flow is not possible in currently known physical implementation, including biology.

\item {\bf Architecture 2:} This represents a physical implementation of BP where the same neurons are used in the forward and learning channels, with a separate identical set of connections, exactly mirroring the forward connections, in the learning channel
(Figure \ref{fig:BP-ARCH123}). This is the Conjoined Case. This implementation uses transpose matrices $A^t$ and in essence corresponds to how BP is viewed when implemented in a digital computer.
Such an implementation in a physical system is faced by major challenges in terms of ARC realization and WTS.
If these were solved, then NEU and DER could also be solved as a byproduct of being conjoined. The ADA problem is addressed only if one can postulate a corresponding mechanism to maintain the weight symmetry at all times during learning. If the WTS problem is solved only at initialization, then the ADA problem is a challenge. Finally, this architecture does not address the LIN problem. 

\item {\bf Architecture 3:} This represents a physical implementation of BP using a set of neurons and connections in the learning channel that is clearly distinct from the neurons and connections in the forward channel(Figure \ref{fig:BP-ARCH123}). This is the Twin Case when the architecture in the learning channel is identical to the forward architecture. This implementation is faced with all six symmetry challenges: ARC, WTS, NEU, DER, ADA, and LIN.
The Identical Twin subcase correspond to having a one-to-one map between neurons in the forward and learning channel, which solves the NEU problem. 
In a digital computer implementation, the Conjoined and Identical Twin Cases are essentially the same.

\item {\bf Architecture 4:} This represents a physical implementation of RBP in the Conjoined Case, using the same neurons in the forward and the learning channel
(Figure \ref{fig:RBP-ARCH45}).
Each forward connection is mirrored by a connection in the reverse direction, but the forward and backward connection have different weights. The weights on the backward connections are random and fixed. This corresponds also to the standard implementation of RBP in a digital computer and, as such, addressed the WTS challenge.
The ARC and NEU symmetries are inherent in the Conjoined architecture, and the DER challenge can be addressed as a byproduct. (Without multiplication by the derivative of the activation functions, RBP does not seem to work.) 
The LIN and ADA challenges are not addressed by standard RBP. Simulations carried in 
\cite{baldiRBP2016AI}, however with no supporting theoretical results, show that if each random weight is adapted proportionally to the product of the forward signal (postsynatpic term in the backward direction) and the randomly backpropagated error (presynaptic term in the backward direction),
then learning converges.

\item {\bf Architecture 5:} This represents a physical implementation of SRBP (skipped RBP) in the Conjoined Case,
using the same neurons in the forward and the learning channel (Figure \ref{fig:RBP-ARCH45}).
Each top neuron (where the error is computed) is connected to each deep neuron.
The weights on the backward connections are random and fixed. This corresponds also to the standard implementation of SRBP in a digital computer
and, as such, addressed the WTS challenge.
The ARC and NEU symmetries are inherent in the Conjoined skipped architecture, and the DER challenge can be addressed as a byproduct. (Without multiplication by the derivative of the activation functions, RBP does not seem to work.) 
Importantly, this implementation shows that when updating a forward weight, only the derivative of the activation of its postsynaptic neuron matters. All other derivatives can be ignored. The LIN and ADA challenges are not addressed by standard SRBP. Simulations carried in 
\cite{baldiRBP2016AI}, however with no supporting theoretical results,  show that if each random weight is adapted proportionally to the product of the forward signal (postsynatpic term in the backward direction) and the randomly backpropagated error (presynaptic term in the backward direction),
then learning converges.

\item {\bf Architecture 6:} This represents a physical implementation of RBP using a set of neurons and connections in the learning channel that is clearly distinct from the neurons and connections in the forward channel
(Figure \ref{fig:RBP-ARCH67}). This is the Twin Case if the architecture is the same in both pathways, and Identical Twin if there is a one-to-one correspondence between the neurons in each pathway.
The WTS challenge is addressed by the random weights. However the NEU and DER challenges remain major challenges, even if the ARC challenge is fully addressed, and so are the LIN and ADA challenges. Nevertheless, simulations carried in \cite{baldiRBP2016AI} show that the random connection can be adapted using the same algorithm described in Architecture 4.

\item {\bf Architecture 7:}
This represents a physical implementation of SRBP. It is identical to Architecture 6, except with skip connections in the learning channel (Figure \ref{fig:RBP-ARCH67}).
This is the Twin Case if the architecture is the same in both pathways, and Identical Twin if there is a one-to-one correspondence between the neurons in each pathway.
The WTS challenge is addressed by the random weights. However the NEU and DER challenges remain major challenges, even if the ARC challenge is fully addressed, and so are the LIN and ADA challenges. Nevertheless, simulations carried in \cite{baldiRBP2016AI} show that the random connections can be adapted using the same algorithm described in Architecture 4.

\item {\bf Architecture 8:} This represents a physical implementation of RBP (or similarly for SRBP) similar to Architecture 6 (Figure \ref{fig:RBP-ARCH8}), corresponding to the Twin Case if the architectures in both pathways are identical. However connections with random weights ( matrices $C'$) are used to addressed the NEU challenge in one direction. The LIN and ADA problems remain as above. 

\item {\bf Architecture 9 and 10:} This represents a physical implementation of RBP (or similarly for SRBP) similar to Architecture 6, however connections with random weights (matrices $C'$ and $A'$) are used to addressed the NEU problem in both directions (Figure \ref{fig:RBP-ARCH9}).
This can be for the Twin Case, or the more general Distinct case, where the architecture of the learning channel is distinct and different (at least in terms of layer sizes) from the forward channel. Figure
\ref{fig:FinalRBP} is simply a variation in which the learning channel has some combination of standard and skip connections.
One goal of this work is to address the LIN and ADA problems in this architecture, by using  the same non-linear neurons in the forward channel and the learning channel, and using the same learning rule--including the derivative of the local activation function--in both channels.
\end{itemize}

In summary, RBP directly solves the symmetry of weights problem (WTS), immediately showing that symmetry is not needed. However the plain RBP algorithm is computed on an architecture that mirrors the forward architecture and thus by itself it does not solve the first symmetry problem (ARC). 
This problem is solved by SRBP and RBP when the learning channel is implemented in a separate architecture (Distinct), which could even include a combination of SRBP and RBP connections.

RBP and SRBP also provide an elegant solution to the third symmetry problem, the correspondence problem (NEU). In particular, the learning channel does not have to know which neuron is which in a given layer of the forward network. It simply connects randomly to all of them. Finally, simulation studies in \cite{baldiRBP2016AI} show that only the derivative of the activation functions of the neuron in layer $h$ whose weights are being updated are necessary, which addressed the transport of derivatives problems (DER). This information is local and information about all the derivatives of the activations in layers above $h$ are not necessary.
Thus we are left essentially with the last two symmetry problems (LIN and ADA). Through simulations we are going to show that it is possible to use the same non-linear neurons in both the forward channel and the learning channel and, in addition, it is possible to let the weights in the learning channel adapt using the same learning rule as the forward weights. We will also be able to prove convergence results when both channels are adaptive, at least in some simplified cases.

\subsection{Other Learning Rules (STDP)}

As previously discussed, in most of the simulations and the mathematical results we use the learning rule:
\be
\Delta w_{ij}=\eta (f^{post})'O_i^{post} O_j^{pre}
\label{eq:}
\ee
for both channels, where here $w_{ij}$ represents the synaptic weight of a directed connection in {\it either} channel and $(f^{post})'$ is the derivative of the activity of the postsynaptic neuron. The presynaptic terms correspond to activity in the same channel as the weight $w_{ij}$, whereas the postsynaptic terms correspond to activity originated in the opposite channel.
This approach requires neurons to be able to make a distinction between signals received from the forward channel and the learning channel and to be able to remember activities across different channel activations.

Other Hebbian or anti-Hebbian learning rules have been proposed, in connection with spike time dependent synaptic plasticity (STDP), based on the temporal derivative of the activity of the postsynaptic neuron \cite{NIPS1999_1658}. These temporal derivatives could be used to encode error derivatives. Within the present framework which uses non-spiking neurons, these learning rules rely on the product of the presynaptic activity times the rate of change of the postsynaptic activity:

\be
\Delta w_{ij}=\eta (\Delta O_i^{post})(O_j^{pre})
\label{eq:STDP1}
\ee
with a negative sign in the anti-Hebbian case. For simplicity, we denote this kind of learning rule as a STDP rule, even if we do not use spiking neurons in this work. 
For a deep weight $w_{ij}^h$, we can write:

\be
\Delta w^h_{ij}=\eta (\Delta O_i^{h})(O_j^{h-1})
\label{eq:}
\ee
To establish a connection to error derivatives, it is easiest to consider the SRBP framework and 
consider that at $t=0$:

\be
O_i^h(t=0)=f_i^h(\sum_jw_{ij}^hO_j^{h-1})
\label{eq:}
\ee
Now consider that at $t=1$ the output is fed back, by the random connections in the learning channel, giving:

\be
O_i^h(t=1)=f_i^h(\sum_jw_{ij}^hO_j^{h-1} + \sum_k 
c^h_{ik}O^L_k)
\label{eq:}
\ee
Finally, consider that at $t=2$ the output is clamped to the target $T$ and fed back by the random connection in the learning channel, giving:

\be
O_i^h(t=1)=f_i^h(\sum_jw_{ij}^hO_j^{h-1}  + \sum_k 
c^h_{ik}T_k)
\label{eq:}
\ee
Then, provided the weights $c$ in the learning channel are small, we have:

\be
\Delta O_i^h =O^h_i(t=2)-O^h_i(t=1) \approx 
f_i^{h}{'}(\sum_jw_{ij}^hO_j^{h-1} )\sum_kc_{ik}^h (T_k-O^L_k)=
R_i^h 
\label{eq:}
\ee
Thus the resulting learning rule in the forward channel is identical or very similar to SRBP. However, in the learning channel, the same reasoning leads to a different learning rule given by

\be
\Delta c_{ik}^h=\eta T_k \Delta O_i^h= \eta T_k R_i^h
\label{eq:}
\ee
A similarly inspired rule can be derived also in the non-skipped case.
Thus, in short,  for completeness we will also present simulation results for this class of STDP learning rules, and derive a proof of convergence in a simple case (see Section 5.2).

\section{Simulations}

In this section, various implementations of the learning channel are investigated through simulations on three benchmark classification tasks: the MNIST handwritten-digit data set~\cite{lecun1998gradient},  synthetic data sets of increasing complexity as in ~\cite{bianchini_2014}, and the HIGGS data set from high-energy physics~\cite{baldi_searching_2014}. We start with the relatively easy MNIST task, then confirm some of the main results using the more difficult synthetic and HIGGS tasks.



\subsection{MNIST Experiments}

On the MNIST task, the forward channel consisted of 784 inputs, four fully-connected hidden layers of 100 tanh units, and a softmax output layer with 10 units. All weights were initialized by sampling from a scaled normal distribution~\cite{glorot_2010}, and bias terms were initialized to zero.  Training was performed for 100 epochs using mini-batches of size 100 with no momentum and a learning rate of $0.1$ unless otherwise specified. Training was performed on 60,000 training examples and tested on 10,000 examples.

\subsubsection{Non-Linearity in the Learning Channel}

The following simulations investigate learning channels made from non-linear processing units. The learning channel is characterized by (1) the type of non-linear transfer function, (2) the architecture (Conjoined or Distinct), and (3) the algorithm (BP, RBP, or SRBP). We focus here on models that use the tanh non-linearity in both the forward channel and the learning channel, but other non-linearities are discussed.

\begin{figure}[H]
\begin{subfigure}[b]{\textwidth}
\includegraphics[width=\textwidth]{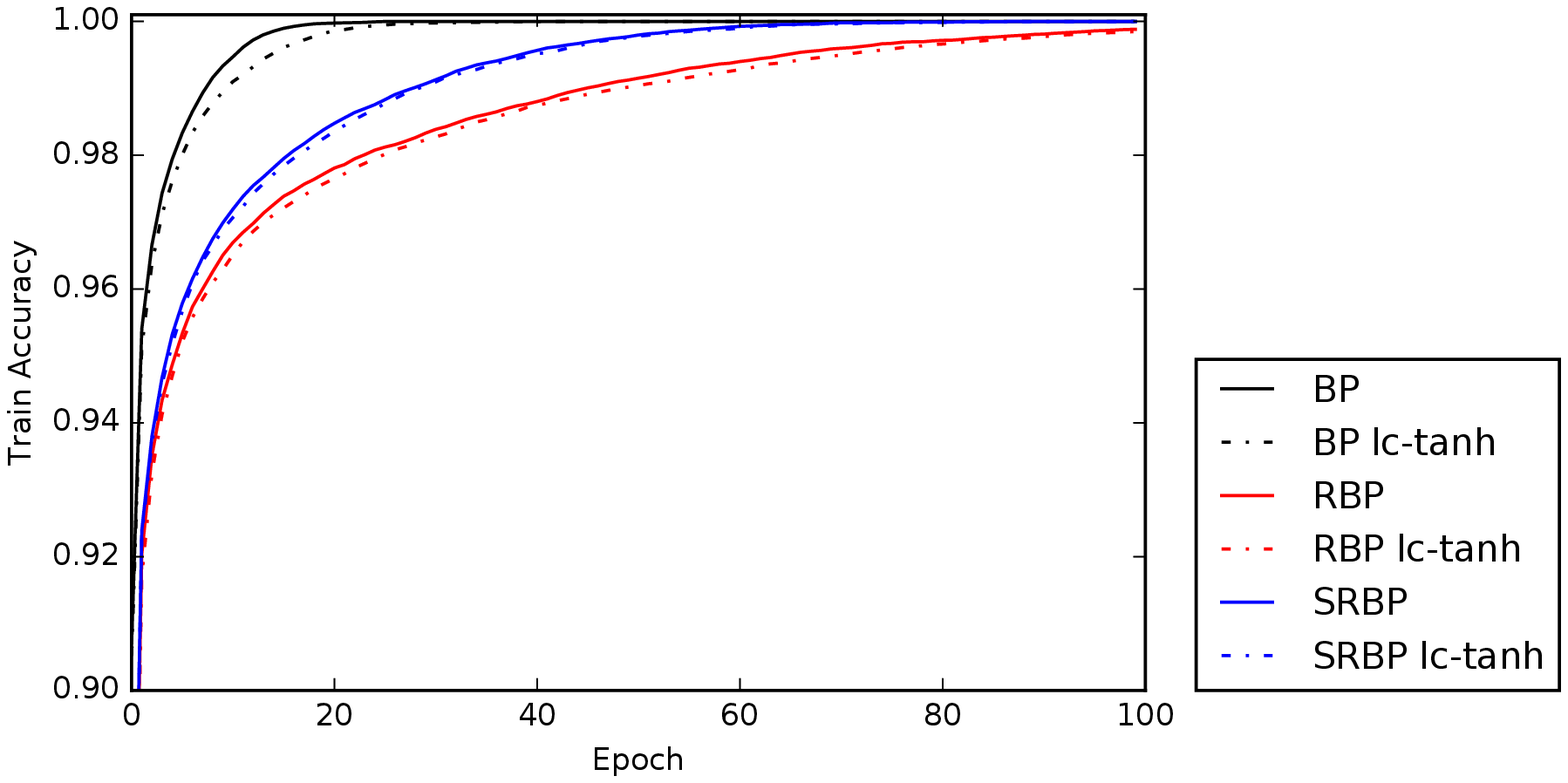}
\end{subfigure}
\begin{subfigure}[b]{\textwidth}
\includegraphics[width=\textwidth]{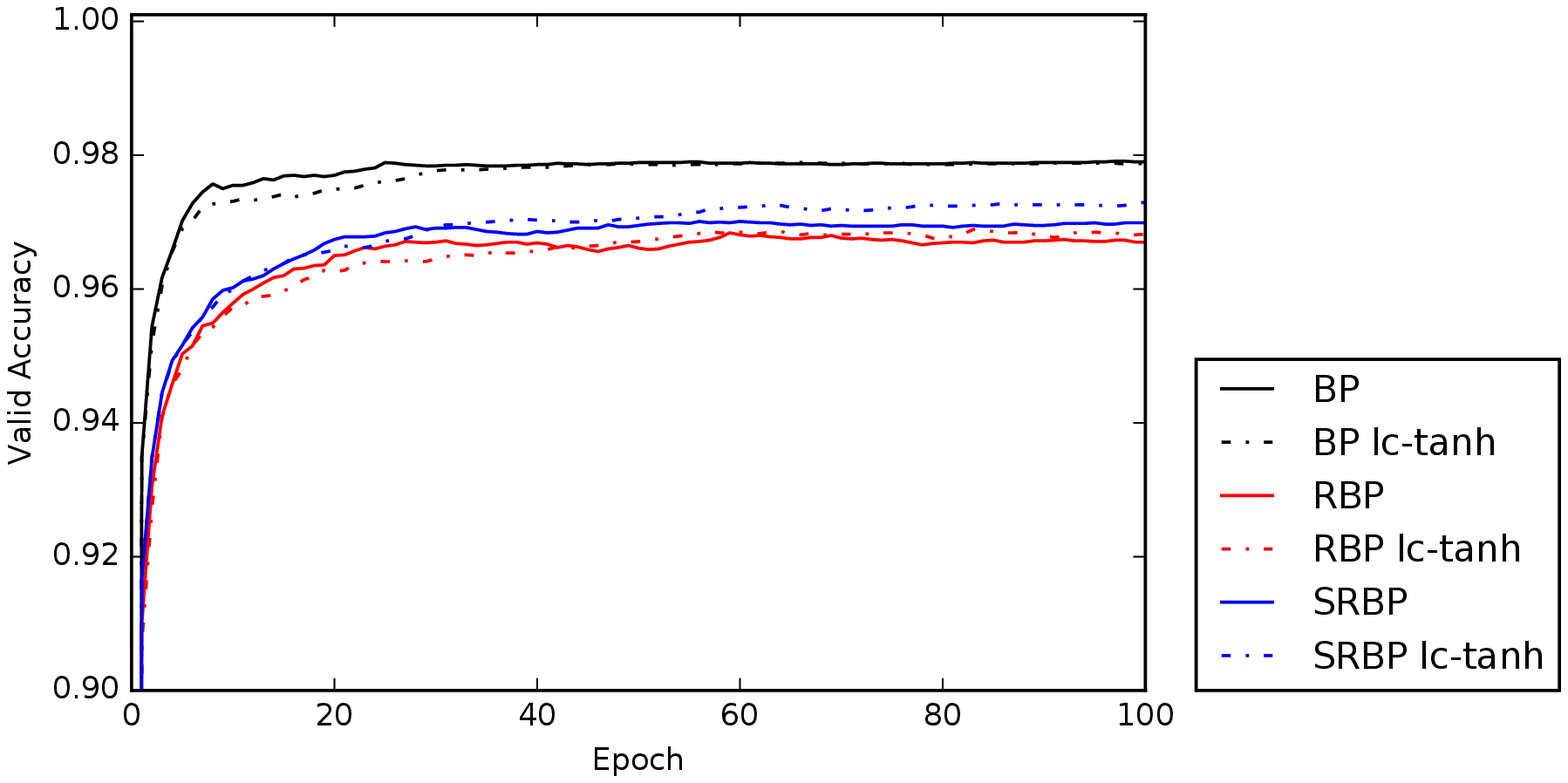}
\end{subfigure}
\caption{MNIST training and validation performance trajectories, as a function of training epoch, for Conjoined architectures, trained with the three algorithms (BP, RBP, SRBP). For each algorithm, we compare the original algorithm to a variant where the tanh non-linearity is applied to the error signal at each neuron (lc-tanh).}
\label{fig:mnist_conjoined}
\end{figure}

In a Conjoined architecture, the error signal is propagated backwards through each neuron in the forward channel, and is modulated by the derivative of the transfer function. Our first simulation examines the effect of applying a non-linearity to the backpropagated error signal summation, immediately before it is multiplied by the transfer function derivative. Figure \ref{fig:mnist_conjoined} shows that the performance of the BP, RBP, and SRBP algorithms does not suffer from this minor modification. Here, BP is represented by Architecture 2 in Figure \ref{fig:BP-ARCH123}, RBP by Architecture 3 in Figure \ref{fig:RBP-ARCH45}, and SRBP by Architecture 4 in Figure \ref{fig:RBP-ARCH45}. 

It should be noted that the non-linearity has little effect when the error signal being propagated through the learning channel is small. So while we verified experimentally that the error signals sometimes fall in the non-linear regime at initialization and early in training, the impact of the non-linearity is small after the network fits the data. We also note that the behavior can be very different for other non-linearities. If the non-linearities in both the forward and the learning channels have a non-negative range, such as the logistic or rectifier functions, then both the neuron activities in the forward channel and the error signals are positive, leading to monotonically increasing weights and poor learning.

\begin{figure}[H]
\begin{subfigure}[b]{\textwidth}
\includegraphics[width=\textwidth]{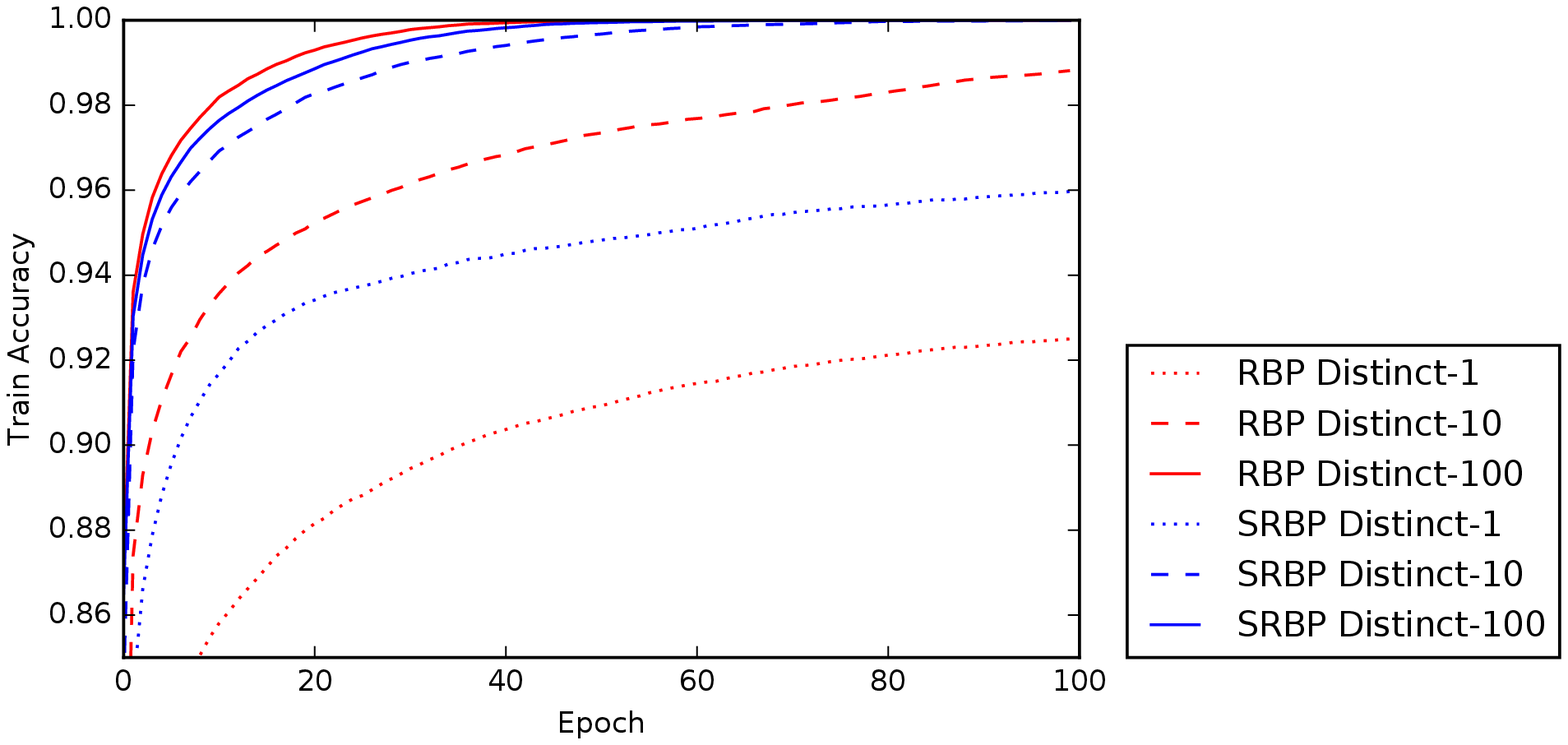}
\end{subfigure}
\begin{subfigure}[b]{\textwidth}
\includegraphics[width=\textwidth]{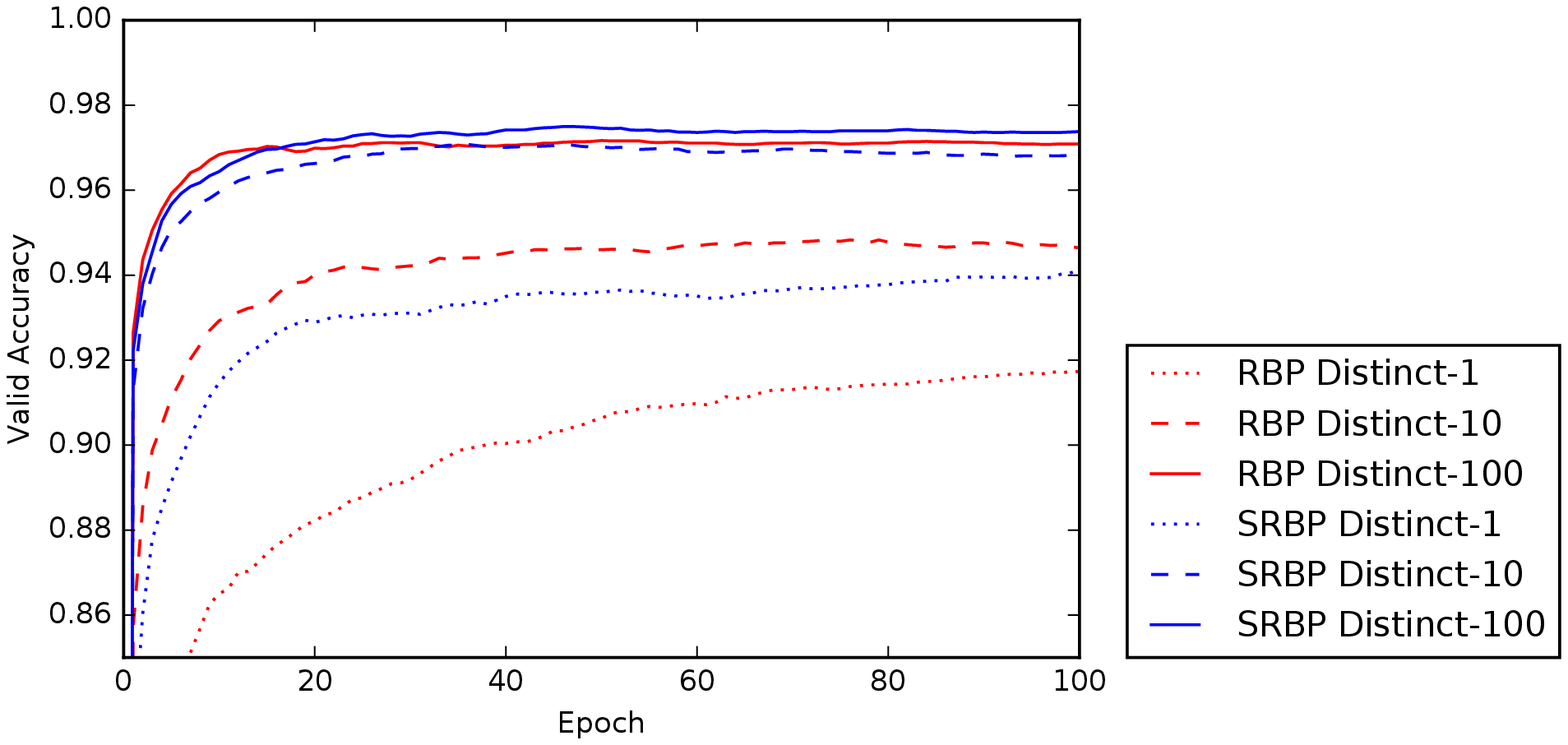}
\end{subfigure}
\caption{MNIST training and validation performance trajectories, as a function of training epoch, for Distinct architectures. The forward channel consists of four hidden layers of 100 tanh neurons, while the learning channel consists of a completely separate set of tanh neurons (with 100, 10, or 1 neuron in each layer) and additional, random, lateral connections that propagate the error signals from the learning channel neurons to the forward channel neurons.}
\label{fig:mnist_distinct}
\end{figure}

In a Distinct architecture, the learning channel consists of a completely separate set of neurons. These learning channel neurons propagate the error signal to the deep layers of the network, then laterally to the corresponding forward neurons via random lateral connections, parameterized by fixed matrices of random weights at each layer (Architecture 8 in Figure \ref{fig:RBP-ARCH8}).  In the SRBP version, skip connections propagate the error signal from the output to each layer of the learning channel, rather than a sequential chain. Figure \ref{fig:mnist_distinct} demonstrates that the model can reach perfect training accuracy with a Distinct architecture. This is true whether the neurons in the learning channel are tanh or linear (not-shown), and whether the learning channel consists of 100 neurons at each layer or 10. A learning channel consisting of a single neuron at each layer leads to slow learning, but still appears to converge.

\subsubsection{Dropout in the Learning Channel}

The dropout algorithm is a common regularization method for neural network models. Here we demonstrate that dropout can also be applied to both the forward channel and the learning channel, where the learning channel consists of tanh neurons organized in a Conjoined or Distinct architecture. During training, the probability of dropping out a hidden neuron in the forward channel is controlled by a parameter $p$, and the activities of neurons that are not dropped are scaled by $1/(1-p)$; dropout in the learning channel is controlled by an analagous parameter $p_{lc}$. At evaluation time, no dropout is used.

\begin{figure}[H]
\begin{subfigure}[b]{\textwidth}
\includegraphics[width=\textwidth]{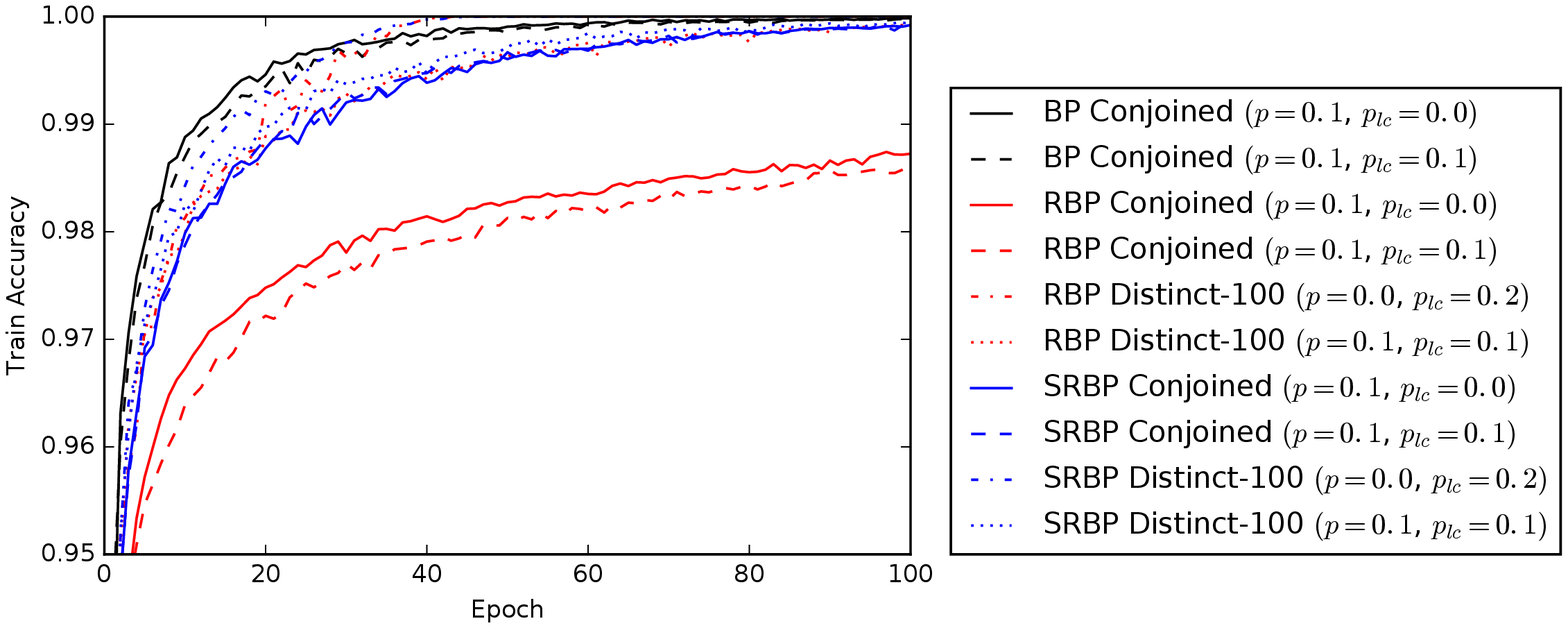}
\end{subfigure}
\begin{subfigure}[b]{\textwidth}
\includegraphics[width=\textwidth]{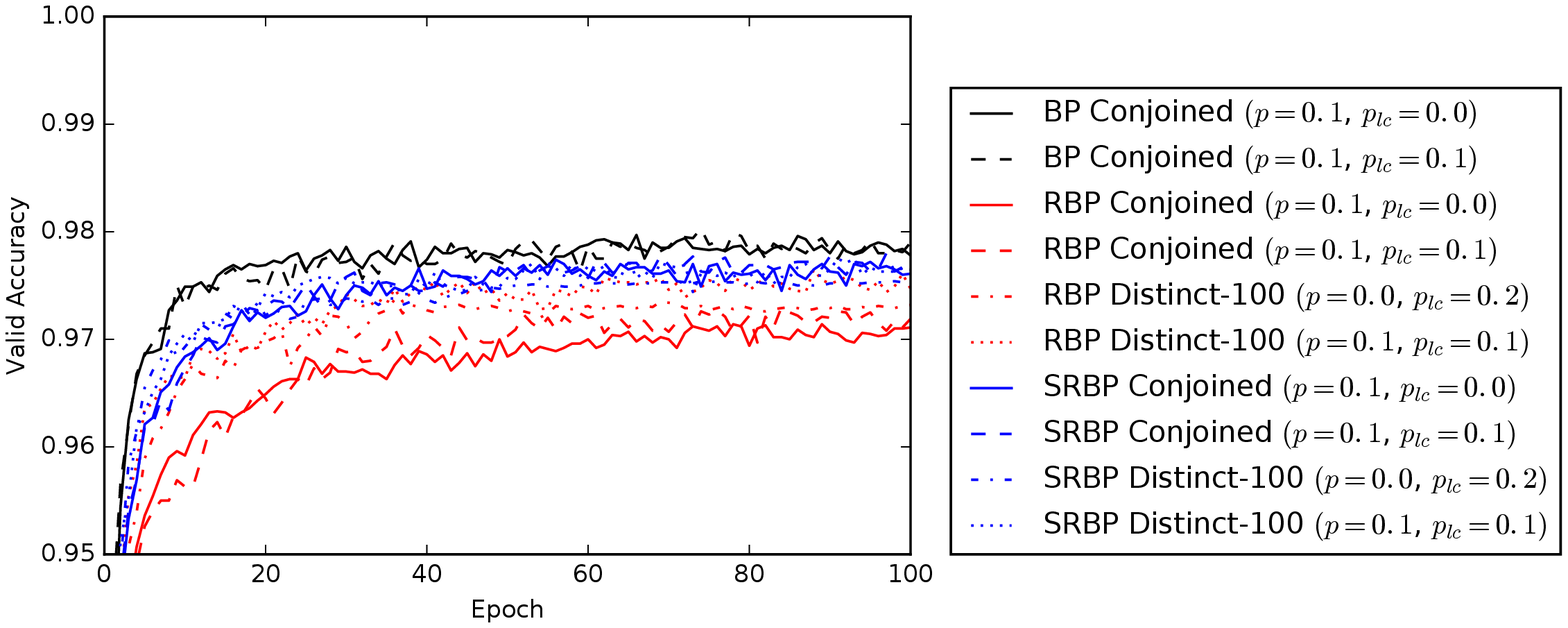}
\end{subfigure}
\caption{MNIST training and validation performance trajectories, as a function of training epoch, for Conjoined and Distinct architectures with a non-linear (tanh) learning channel trained with the three algorithms (BP, RBP, SRBP) and dropout. Dropout is applied to every layer with probability $p$ in the forward channel and probability $p_{lc}$ in the learning channel. The performance of the classifier was evaluated without dropout on the training and test set after every epoch.}
\label{fig:mnist_dropout}
\end{figure}

Figure \ref{fig:mnist_dropout} demonstrates the use of dropout on non-linear (tanh) learning channels in both Conjoined and Distinct architectures. Dropout is independently applied to all hidden layers in both the forward channel and the learning channel. As expected, dropout in the forward channel slowed learning, especially in RBP with a Conjoined architecture. However, the effect of dropout in the learning channel was small regardless of algorithm or architecture. From these results, it cannot be said whether dropout in the learning channel contributes to regularization, but it appears to interfere with learning less than dropout in the forward channel.

\subsubsection{Adaptation in the Learning Channel}

In the simulations so far, randomly-initialized parameters in the learning channel remain constant while the parameters in the forward channel are trained. In this section, we investigate adaptive learning channels where the parameters are randomly-initialized but are updated during training according to the local learning rules defined in Section 2.

\begin{figure}[H]
\begin{subfigure}[b]{\textwidth}
\includegraphics[width=\textwidth]{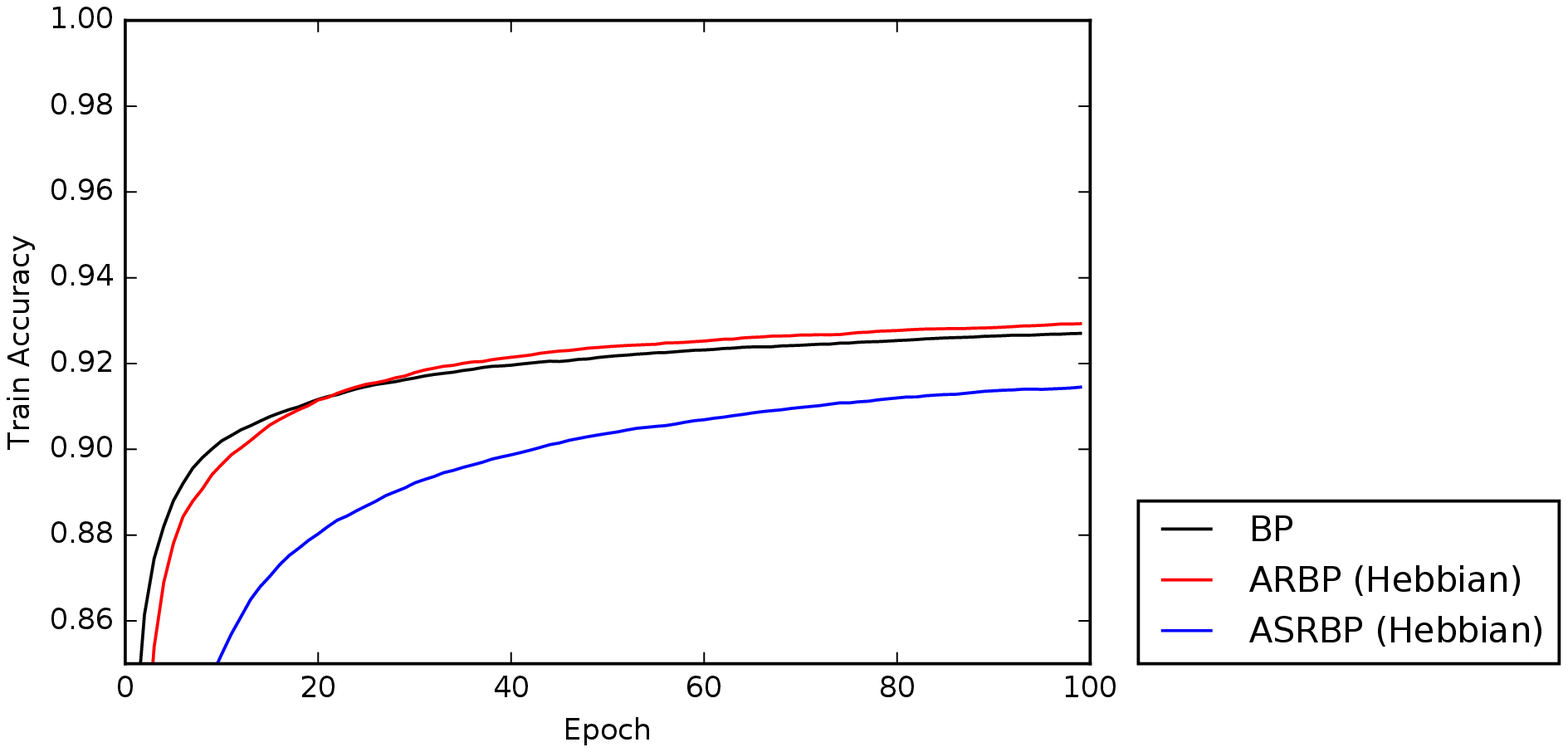}
\end{subfigure}
\begin{subfigure}[b]{\textwidth}
\includegraphics[width=\textwidth]{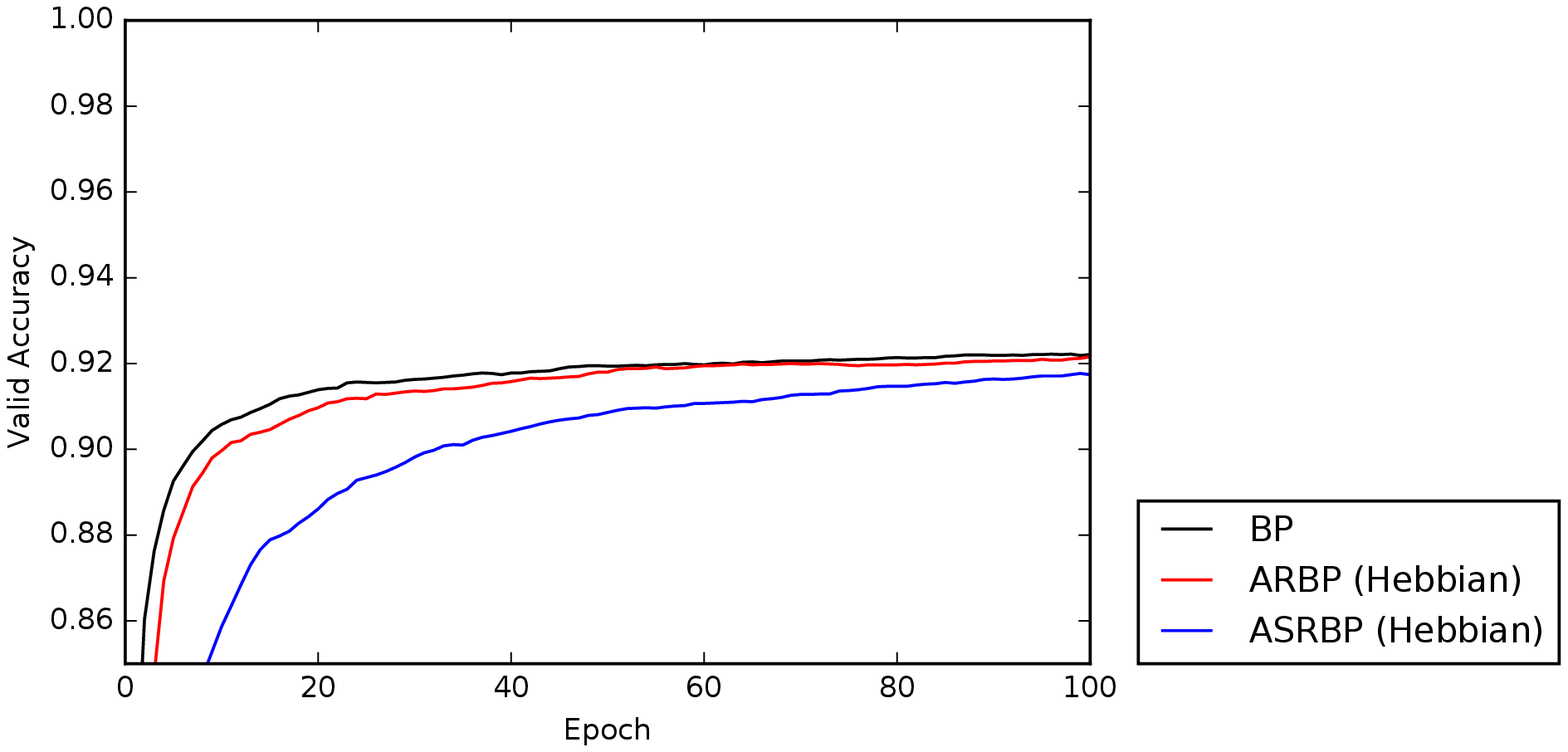}
\end{subfigure}
\caption{MNIST training and validation performance trajectories, as a function of training epoch, for a linear architecture (Conjoined) trained with standard backpropagation (BP), and the adaptive versions of RBP and SRBP with the Hebbian rule (ARBP and ASRBP). Performance does not reach 100\% training accuracy because of the limitations of a linear architecture. The results demonstrate how Hebbian adaptation in deep linear networks leads to performance that is similar to BP.}
\label{fig:mnist_adaptive_linear}
\end{figure}

First we examine the Hebbian adaptive rule. Mathematical analysis on simple architectures suggest that the Hebbian adaptive versions of ARBP and ASRBP could converge to a minimum error solution for deep, linear, Conjoined architectures. Figure \ref{fig:mnist_adaptive_linear} demonstrates this behavior on a MNIST classifier with four hidden layers of linear neurons and a softmax output (no non-linearity is used in the learning channel either). In the case of ARBP, the intuition for why this works is that the learning channel matrices at each layer are updated in the same direction as the forward channel matrices, so after training they are approximately transposes of one another, and thus ARBP approximates BP.

\begin{figure}[H]
\begin{subfigure}[b]{\textwidth}
\includegraphics[width=\textwidth]{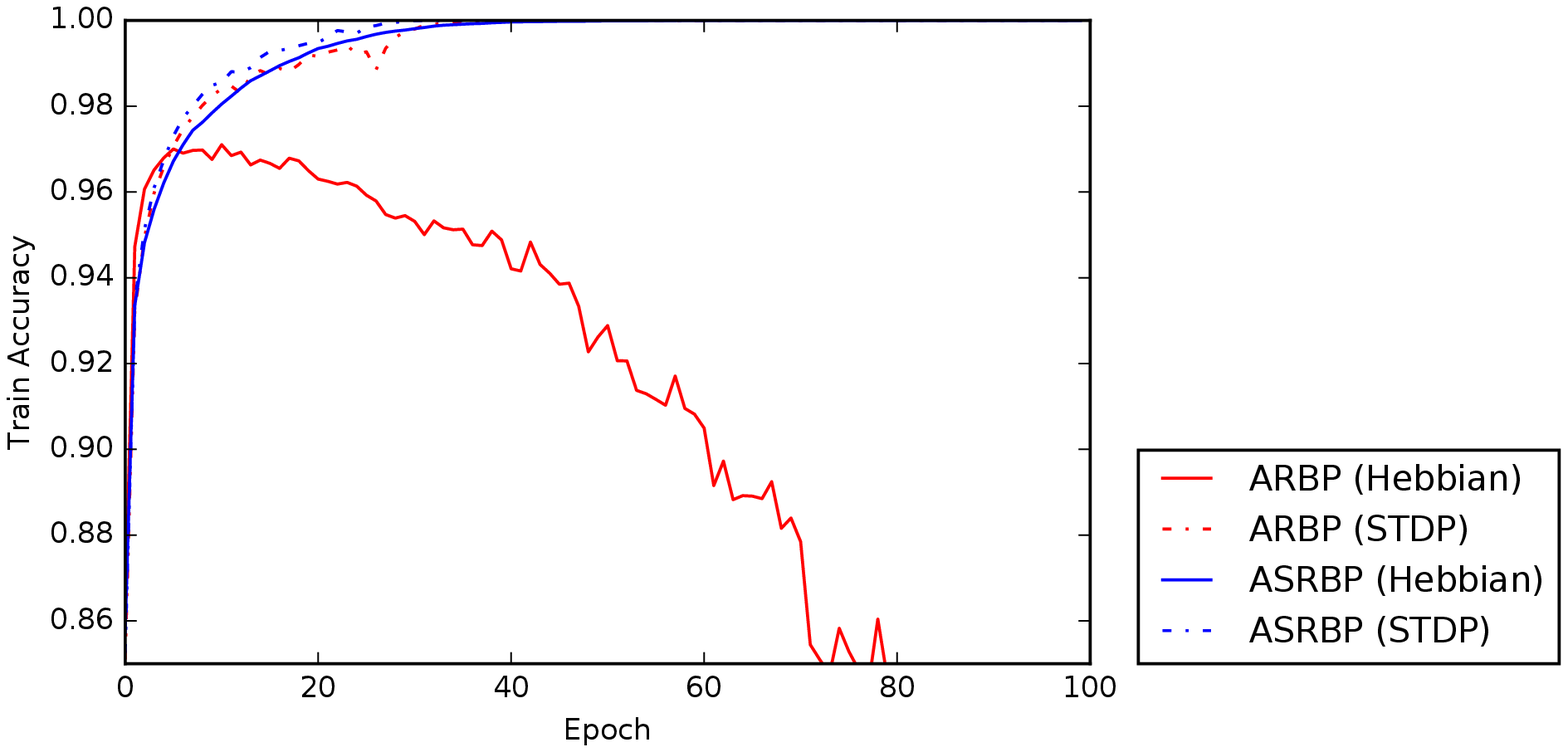}
\end{subfigure}
\begin{subfigure}[b]{\textwidth}
\includegraphics[width=\textwidth]{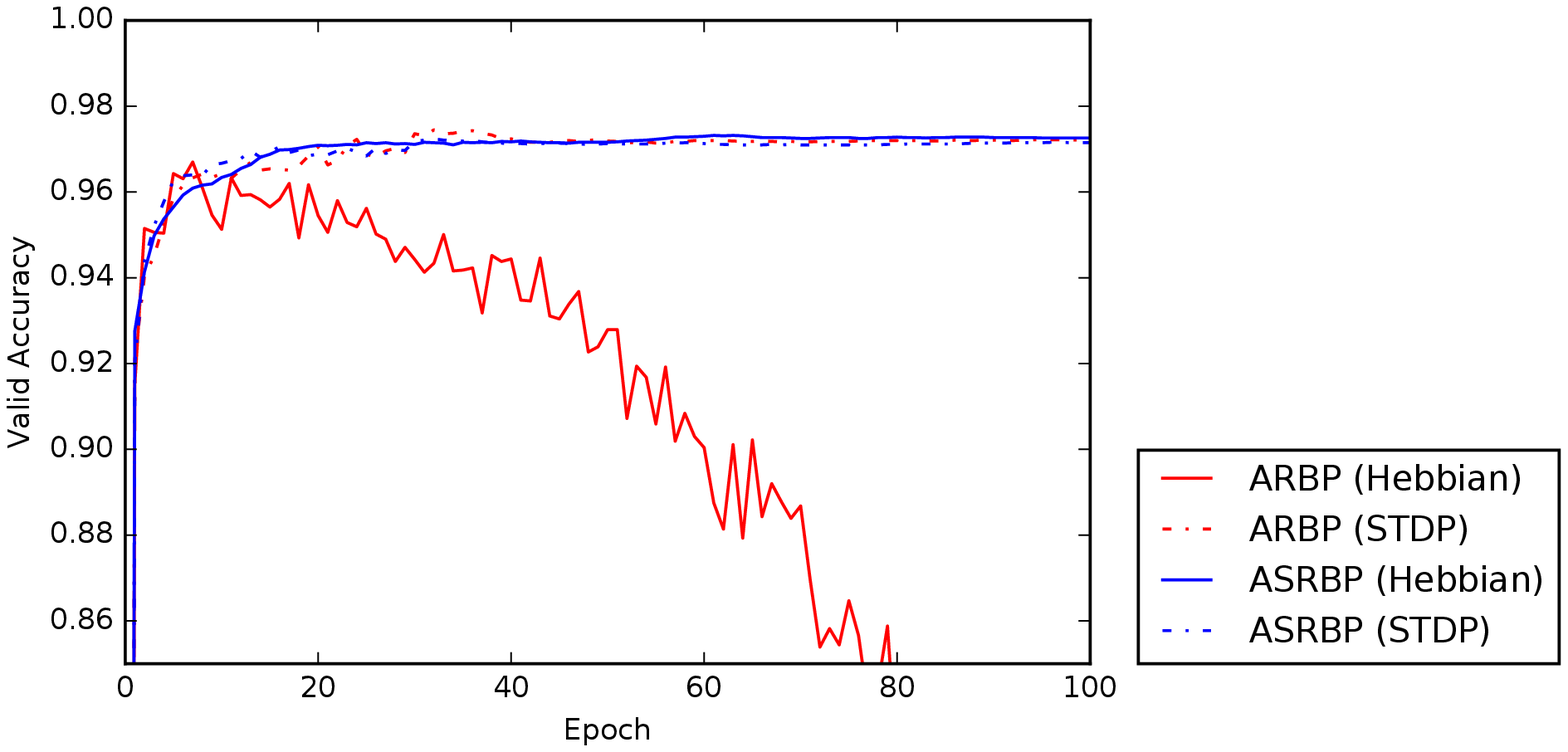}
\end{subfigure}
\caption{MNIST training and validation performance trajectories, as a function of training epoch, for a Conjoined architecture with tanh  units in the forward channel and linear adaptive units in the learning channel. The learning channel weights are updated according to either the Hebbian rule or the STDP rule.}
\label{fig:mnist_adaptive_conjoined}
\end{figure}

The situation becomes more complicated with non-linearities, and our experiments demonstrate that adaptation in the learning channel sometimes prevents the system from converging to the minimum error solution. Figure \ref{fig:mnist_adaptive_conjoined} shows the performance of the two adaptive rules (Hebbian and STDP) on a Conjoined architecture with tanh units in the forward channel. The Hebbian ARBP algorithm initially learns quickly, but the weights in the learning channel grow faster than the weights in the forward channel, causing the activities in the forward channel to saturate and leading to a poor solution. This occures even when the tanh non-linearity is used in the learning channel, and the learning channel weight updates are modulated by the derivative of that transfer function (not shown). In Hebbian ASRBP and the STDP adaptive rule, this problem is avoided and the classifier reaches 100\% training accuracy. 


In these experiments, we have investigated each of the six symmetries described in Section 2. The main conclusion is that the learning channel can be implemented in a number of ways that make use of random connections to transmit the error signals. In particular, the learning channel could be physically separate from the forward channel, have a distinct architecture, and could consist of non-linear, adaptive processing units. Next, we confirm these results on a more difficult classification task.

\subsection{Synthetic Data Experiments} 

\begin{figure}[H]
  \centering
  \includegraphics[width=.95\textwidth]{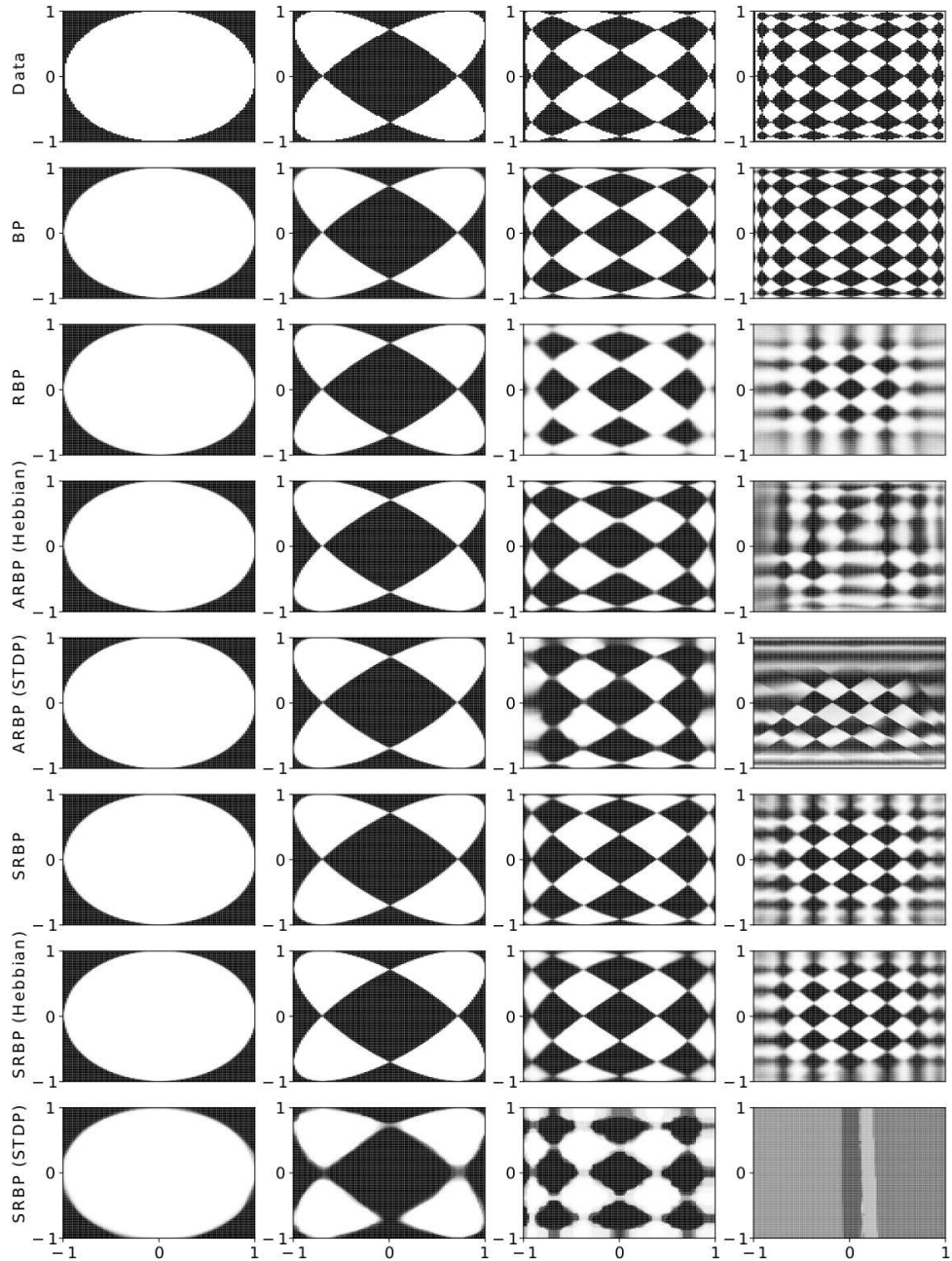}
  \caption{Synthetic classification data sets associated with the functions $f_k = g \circ t_k$, with $g(\mathbf{x}) = \text{sign}(1 - ||\mathbf{x}||_2^2)$, $t(\mathbf{x}) = [1-2x_1^2, 1-2x_2^2]$, $t_0(\mathbf{x}) = \mathbf{x}$, $t_1=t$, $t_2 = t \circ t$, $t_3 = t \circ t \circ t$, etc. Row 1 visualizes these functions and how their complexity increases with $k$, for $k={0,1,2,3}$. Subsequent rows show probabilistic predictions of a fixed neural network architecture trained with the adaptive and non-adaptive versions of RBP and SRBP. With additional hyperparameter tuning (not shown), all the algorithms are able to learn all the functions.}
\label{fig:syntheticdecision}
\end{figure}

Bianchini et al. ~\cite{bianchini_2014} suggest a class of easily-visualized target functions in which the difficulty of the learning problem is parameterized by an integer $k$ that controls the topology of the decision boundaries. 
Specifically, they propose a recursively defined sequence of functions
$f_k: \mathbb{R}^2 \rightarrow \{0,1\}$ defined by:
$f_k = g \circ t_k$, with $g(\mathbf{x}) = \text{sign}(1 - ||\mathbf{x}||_2^2)$, $t(\mathbf{x}) = [1-2x_1^2, 1-2x_2^2]$, $t_0(\mathbf{x}) = \mathbf{x}$, $t_1=t$, $t_2 = t \circ t$, $t_3 = t \circ t \circ t$, etc. 
As illustrated in the top row of Figure \ref{fig:syntheticdecision} by plotting for 
$k={0,1,2,3}$, the functions $f_k$ can be visualized as increasingly complex patterns of white and black regions in the two-dimensional plane. The first two Betti numbers of these functions are $b_0=1$ ($b_0$ is the number of connected components) and $b_1=O(4^k)$ ($b_1$ is the number of one-dimensional or ``circular'' holes). Thus, functions with larger values of $k$ have more holes.

Each function $f_k$ induces a classification learning problem with two-dimensional inputs $(x,y)$ and corresponding targets $f_k(x,y)$. Training examples are generated by sampling $(x,y)$.
Figure \ref{fig:syntheticdecision} shows $f_k$ for $k\in {0,1,2,3}$, as well as predictions of neural networks trained with the different learning rules. The same Conjoined neural network architecture was trained with every algorithm, and consisted of five layers of 500 hidden units per layer, followed by a single logistic output unit. Weights were initialized by sampling from a scaled uniform distribution~\cite{glorot_2010}, and bias terms were initialized to zero. Training was performed on mini-batches of 100 random examples, using stochastic gradient descent without momentum. The learning rate was initialized to 0.01 and decayed by a factor of $10^{-5}$ after each weight update. Training was stopped after 1.5 million iterations, or when the validation error increased by more than 1\%  over a 5000-iteration epoch. The only hyperparameter that was not constant across algorithms was the non-linear transfer function, which was $max(0,x)$ (ReLU) for all but the ARBP (STDP) and ASRBP (STDP) algorithms, which had better performance with tanh neurons. 

The easier data sets (k=0 and k=1) are learned by all the algorithms with a high degree of accuracy. On the more difficult data sets (k=2 and k=3), the random backpropagation algorithms perform slightly worse than standard backpropagation for the fixed architecture and hyperparameters used here. However, all the algorithms learn these functions to a high degree of accuracy with additional hyperparameter tuning (not shown).

\subsection{HIGGS Experiments}

The HIGGS data set is a two-class classification task from high-energy physics~\cite{baldi_searching_2014}. Deep learning provides a significant boost in performance over shallow neural network architectures on this task, especially when the input is restricted to 21 low-level features. In the following experiments, the forward channel consists of the 21 low-level inputs, eight fully-connected hidden layers of 300 tanh units, and a single logistic output unit. Weights were initialized by sampling from a scaled normal distribution~\cite{glorot_2010}, and bias terms were initialized to zero. Training was performed for 100 epochs using mini-batches of size 100, a momentum factor of 0.9, and a learning rate of $0.1$ unless otherwise specified. Classifiers were trained on 10,000,000 examples and tested on 100,000 examples. 

\begin{figure}[H]
\begin{subfigure}[b]{\textwidth}
\includegraphics[width=\textwidth]{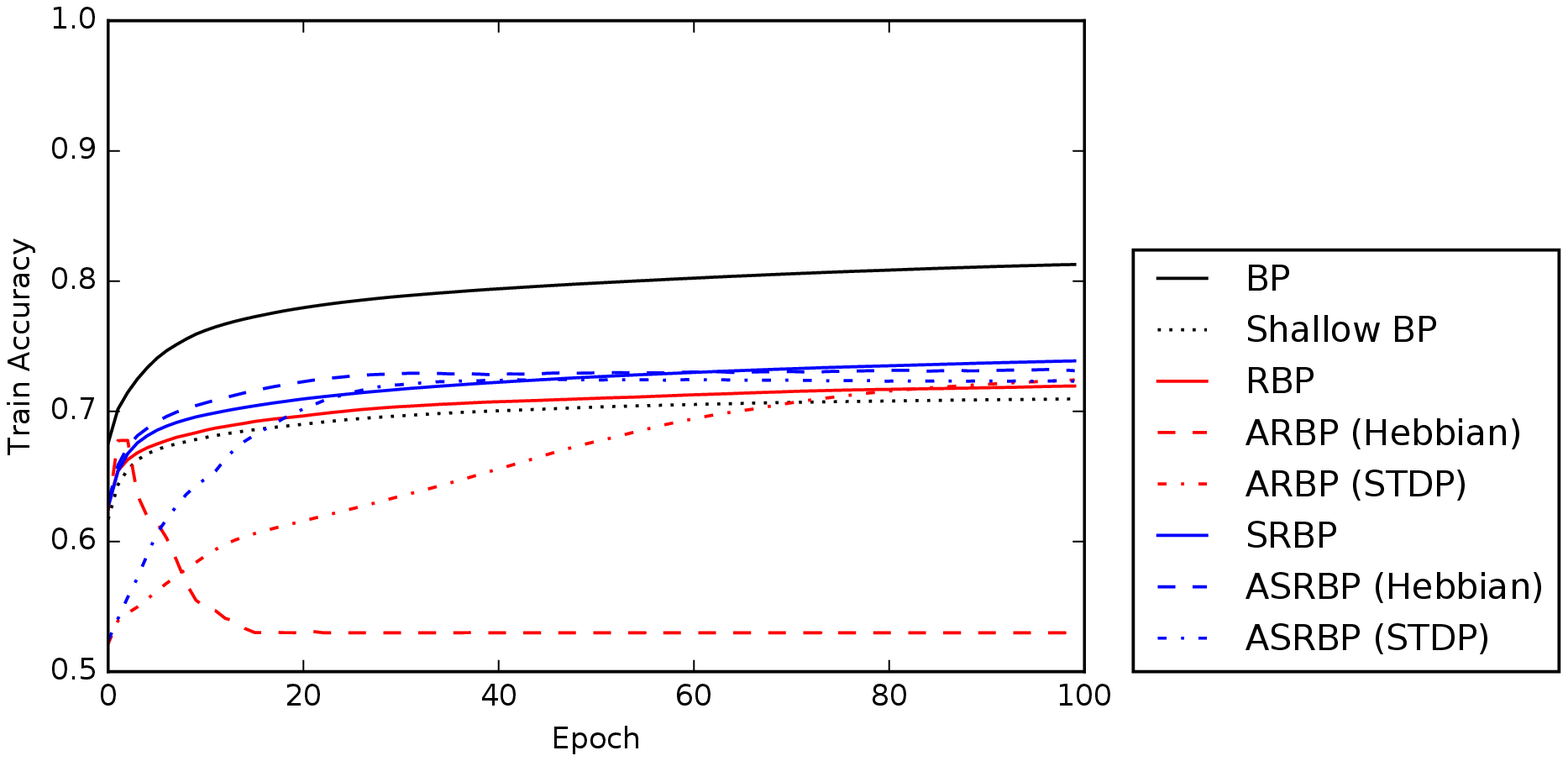}
\end{subfigure}
\begin{subfigure}[b]{\textwidth}
\includegraphics[width=\textwidth]{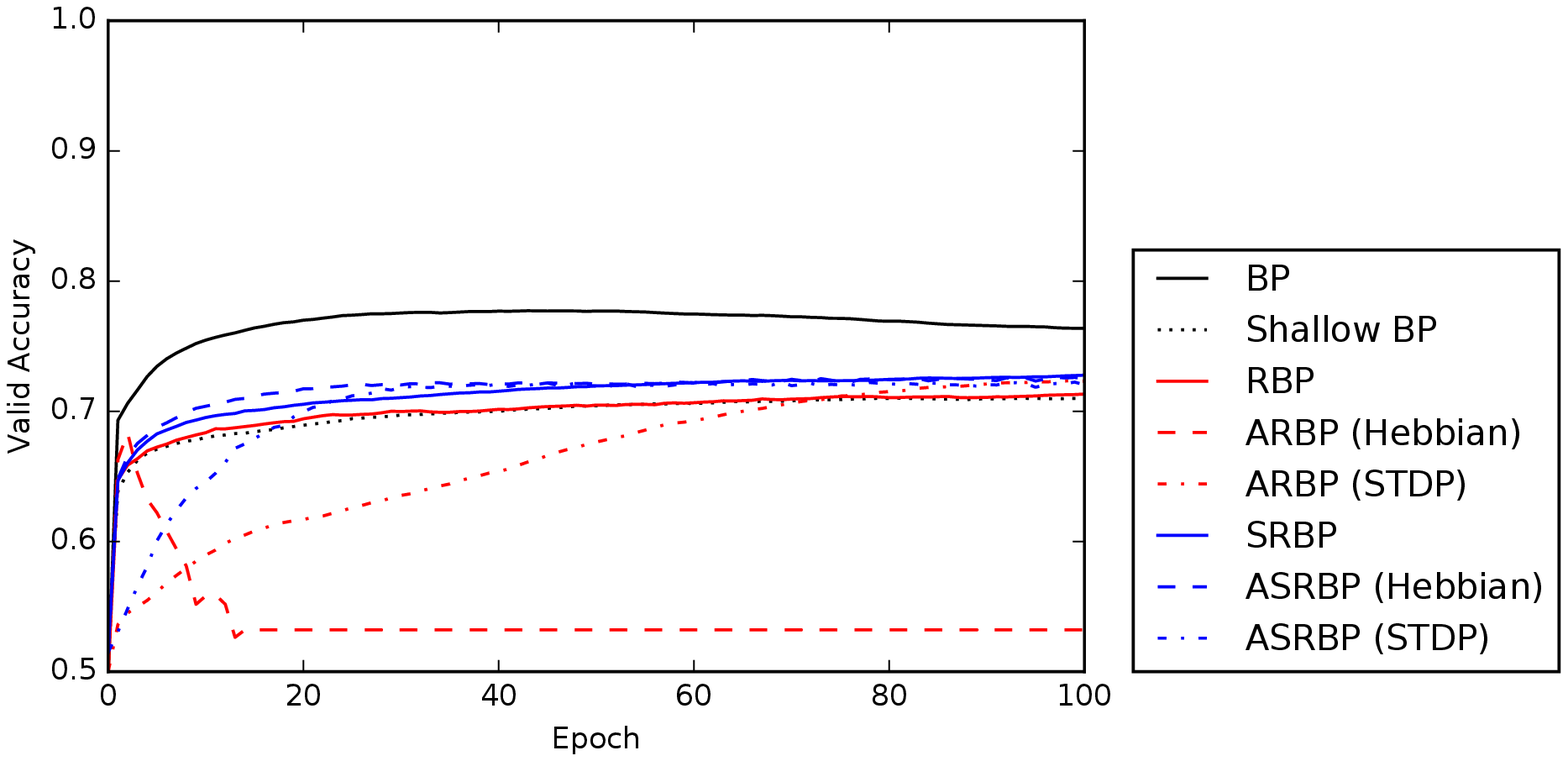}
\end{subfigure}
\caption{HIGGS training and validation performance trajectories, as a function of training epoch, for a Conjoined architecture with tanh neurons in the forward channel and linear neurons in the learning channel. The original BP, RBP, and SRBP are shown along with the adaptive variants of RBP and SRBP (Hebbian and STDP). The STDP variants train slower because they were trained with a smaller learning rate ($0.0003$ rather than $0.1$). As a benchmark, a shallow network consisting of a single hidden layer and trained with BP is also shown.} 
\label{fig:higgs_conjoined}
\end{figure}

The results on Conjoined architectures agree with the results on MNIST. First, we verified that the use of the tanh non-linearity in the learning channel has minimal effect on the performance of BP, RBP, and SRBP (not shown). Figure \ref{fig:higgs_conjoined} shows the results of BP, RBP, and SRBP along with the adaptive variants in Conjoined architectures. As on MNIST, the Hebbian ARBP algorithm learns quickly then decreases in accuracy as the learning channel weights grow too large. However, the other adaptive algorithms perform similarly to their non-adaptive variants, and perform better than a benchmark shallow neural network trained with BP.

\begin{figure}[H]
\begin{subfigure}[b]{\textwidth}
\includegraphics[width=\textwidth]{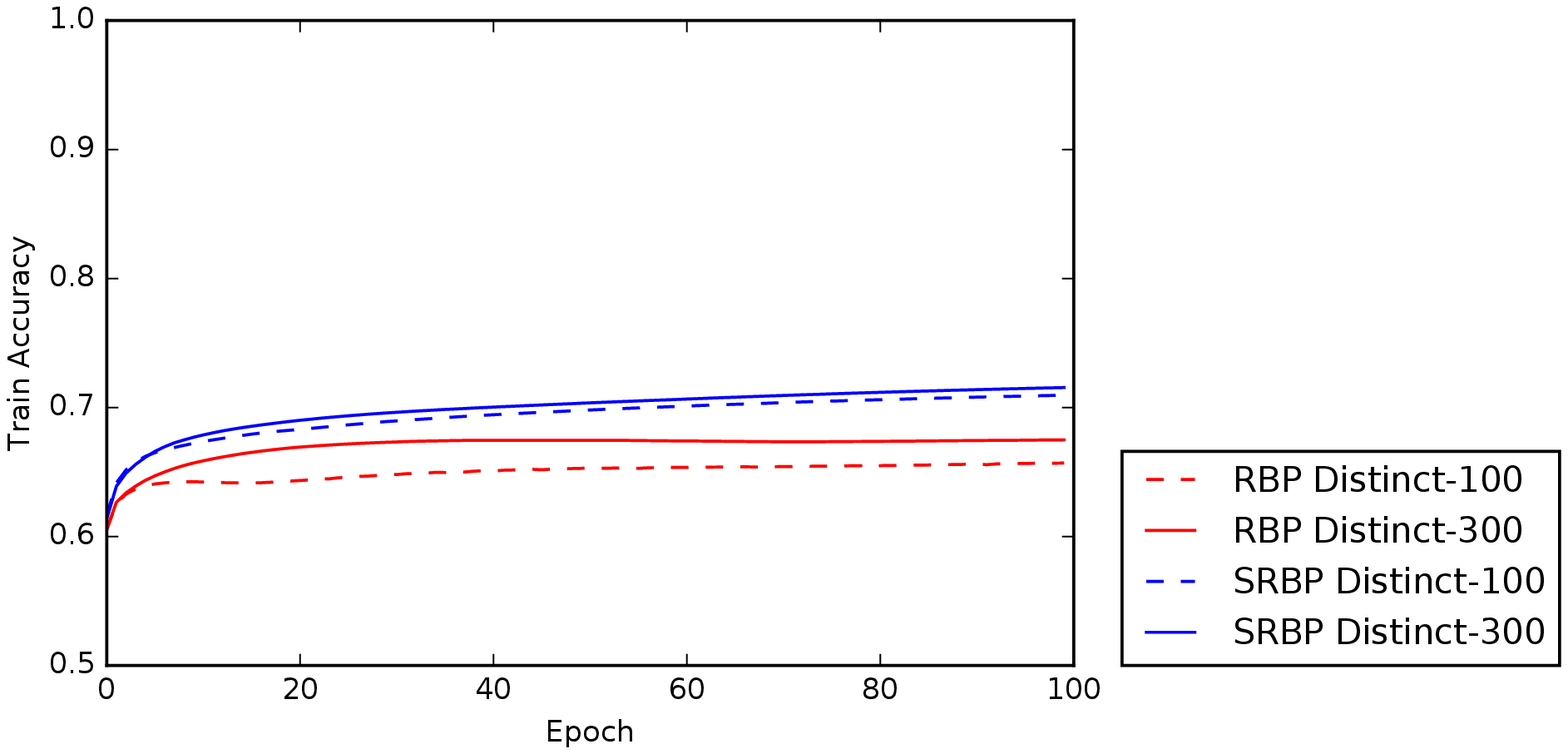}
\end{subfigure}
\begin{subfigure}[b]{\textwidth}
\includegraphics[width=\textwidth]{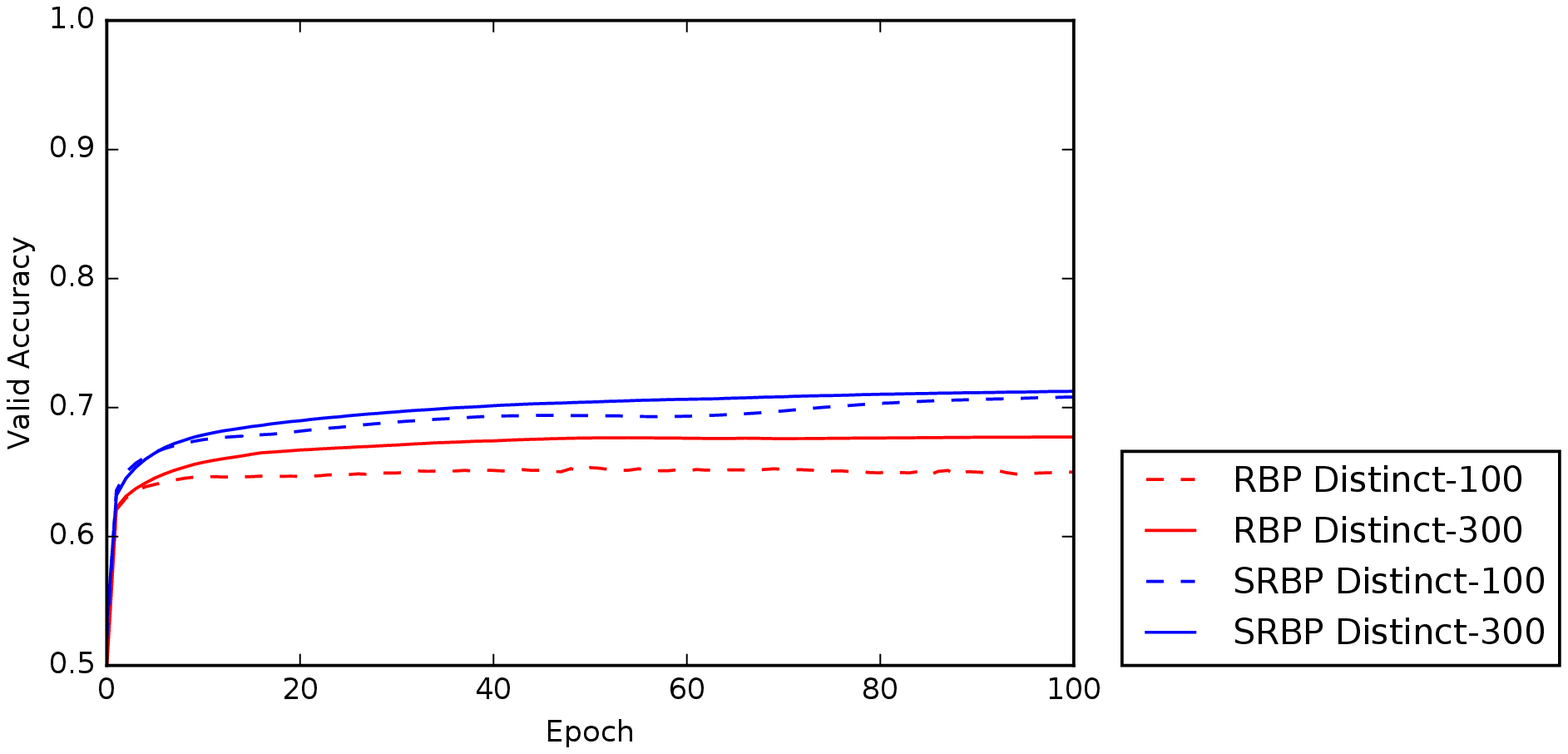}
\end{subfigure}
\caption{HIGGS training and validation performance trajectories, as a function of training epoch, for a Distinct architecture with tanh neurons in both the forward channel and the learning channel. The number of neurons in each hidden layer of the forward channel is 300, while the number of neurons in the learning channel is either 300 or 100. In these experiments, the weights in the learning channel were initialized from a normal distribution with a slightly smaller standard deviation of $\frac{1}{fanin+fanout}$ compared to the $\frac{2}{fanin+fanout}$ used in the forward layer.} 
\label{fig:higgs_distinct}
\end{figure}

The results on Distinct architectures with tanh neurons in both channels are shown in Figure \ref{fig:higgs_distinct}. When the learning channel contains the same number of neurons as the forward channel, SRBP does just as well as in the Conjoined architecture, while RBP does slightly worse. As in the MNIST experiments, changing the number of neurons in the learning channel does not have a large effect on performance.

\section{Mathematical Analyses}

\subsection{General Considerations}

The general strategy to try to derive more precise mathematical results is to proceed from simple architectures to more complex architectures, and from the linear case to the non-linear case. 
In the case of linear networks, when there is no adaptation in the learning channel, then RBP and SRBP are equivalent when
there is only one hidden layer, or when all the layers have the same size. 
However when there is adaptation in the learning channel, then ARBP and ASRBP  are equivalent when there is a single hidden layer, but not when there are multiple hidden layers, even if they are of the same size.
When the learning channel is not adaptive, the differential equations for several kinds of networks were studied in \cite{baldiRBP2016AI}. Here we consider the case of adaptive learning channels. 

For each linear network, under a set of standard assumptions, one can derive a set of non-linear--in fact polynomial--autonomous (independent of time), first order, ordinary differential equations (ODEs) for the average (or batch) time evolution of the synaptic weights under the ARBP or ASRBP algorithm. 
As soon as there is more than one variable and the polynomial system is non-linear, there is no general theory to understand the corresponding behavior. In fact, even in two dimensions,
the problem of understanding the upper bound on the number and relative position of the limit cycles of a system of the form $dx/dt=P(x,y)$ and $dy/dt=Q(x,y)$, where $P$ and $Q$ are polynomials of degree $n$ is open--this is Hilbert's 16-th problem in the field of dynamical systems.

When considering the specific systems arising from the ARBP/ASRBP learning equations, one must first prove that these systems have a long-term solution. Note that polynomial ODEs may not have long-term solutions in general (e.g. $dx/dt=x^\alpha$, with $x(0)> 0$, does not have long-term solutions for $\alpha>1$) but, if the trajectories are bounded, then long-term solutions exist. We are particularly interested in long-term solutions that converge to a fixed point, as opposed to limit cycles or other behaviors. 

A number of interesting cases can be reduced to a first-order, autonomous, differential equation in one dimension $dx/dt=f(x)$ for which long-term existence and convergence to fixed point theorems can be derived. In general we will assume that $f$ is locally Lipschitz over the domain of interest, that is for every $x$ in the domain there is a neighborhood $(x-\epsilon,x+\epsilon)$ such that for any pair of points $(x_1$,$x_2)$ in this neighborhood the function $f$ satisfies the Lipschitz condition:

\be
\vert f(x_2)-f(x_1) \vert \leq K \vert x_2-x_1 \vert
\label{eq:}
\ee
for some constant $K$. The local Lipschitz condition implies that $f$ is continuous, but not necessarily differentiable at $x$. On the other hand, if $f$ is differentiable at $x$ then it is locally Lipschitz.

The fundamental theorem of ordinary differential equations states that if $f$ is locally Lipschitz around an initial condition of the form  $x(t_0)=x_0$, then for some value $\epsilon>0$ there exists a unique solution $x(t)$ to the initial value problem on the interval $[t_0-\epsilon,t_0+\epsilon]$. If $r$ is a fixed point, 
i.e. $r$ is a root of $f$ ($f(r)=0$), and
$f$ is locally Lipschitz over an entire neighborhood of $r$, then the qualitative behavior of the trajectories of the differential equation $dx/dt=f(x)$ with a starting point near $r$ can easily be understood simply by inspecting the sign of $f$ around $r$.
We give two slightly different versions of a resulting theorem that will be used in the following analyses.

\null\par
\noindent
{\bf Theorem 1:}
Let $dx/dt=f(x)$ and assume $f$ is a continuously differentiable function defined over an open, closed, or semi-open interval of the real line  $(a,b)$  where $a$ or $b$ can be finite or infinite. Assume that $f$ has a finite number of roots
$r_i$ ($i=1, \ldots, n$) with $a  \leq r_1 < r_2 \ldots <r_n \leq  b$.  Then for any starting point $x(0)$ in $[r_1,r_n]$ the trajectory converges to one of the roots. The result remains true for any starting point in $(a,r_1]$ provided
$f(a) >0$.
Likewise the result remains true for any starting point in $[r_n,b)$ provided $f(b)<0$. If the domain of $f$ consists of multiple disjoint intervals, then the theorem can be applied to each interval. Furthermore, for any root $r$ of $f$ (fixed point),  its stability is determined immediately by inspecting the sign of $f$ to the left and right of the root. In particular,  
$++$ correspond to attractor on the left, unstable on the right;
$--$ correspond to unstable on the left, attractor on the right;
$+-$ correspond to attractor; and $-+$ correspond to unstable. Finally, if $f$ is a polynomial of odd degree with leading negative coefficient, then $f$ satisfies all the conditions above and $x(t)$ always converges to a fixed point.

\par\null
\noindent
{\bf Theorem 1:}
We consider the extended real line\footnote{with the base of the topology given by the open intervals $(a,b)$ where $a,b\in\mathbb R$ and $(a,+\infty)\cup\{+\infty\}$, and $(-\infty, b)\cup\{-\infty\}$.}  $\hat {\mathbb R}=\mathbb R\cup\{+\infty\}\cup\{-\infty\}$. Let:

\be
a_1<b_1<a_2<b_2<\cdots<a_k<b_k
\label{eq:}
\ee
where all $a_i,b_i\in\hat R$. 
Let $dx/dt=f(x)$ be a first order differential equation in one dimension, where $f(x)$ is locally Lipschitz on $\bigcup_{i=1}^k (a_i, b_i)$. We assume that 
\begin{enumerate}
\item in a neighborhood of $a_i$, $f(x)>0$;
\item in a neighborhood of $b_i$, $f(x)<0$.
\end{enumerate}
Then for any initial value $x(0)\in \bigcup_{i=1}^k (a_i, b_i) $, the system has a long-term solution and is convergent to one of the roots of $f$. Note that the local Lipschitz condition is automatically satisfied when $f$ is continuously differentiable on the real line. Furthermore the theorem is valid even when $f$ has infinitely many roots.

\par \null
\noindent
{\bf Proof:} The proof of either version of this theorem is easily derived from the fundamental theorem of ODE and can easily be visualized by plotting the function $f$.

Finally, in terms of notations, the matrices in the forward channel are denoted by $A_1, A_2, \dots$, and the matrices in the learning channel are denoted by $C_1, C_2, \ldots$
Theorems are stated in concise form and additional important facts are contained in the proofs.

\subsection{The Simplest Linear Chain:  ${\cal A}[1,1,1]$}

\begin{figure}[h!]
    \centering
    \includegraphics[width=0.95\textwidth]{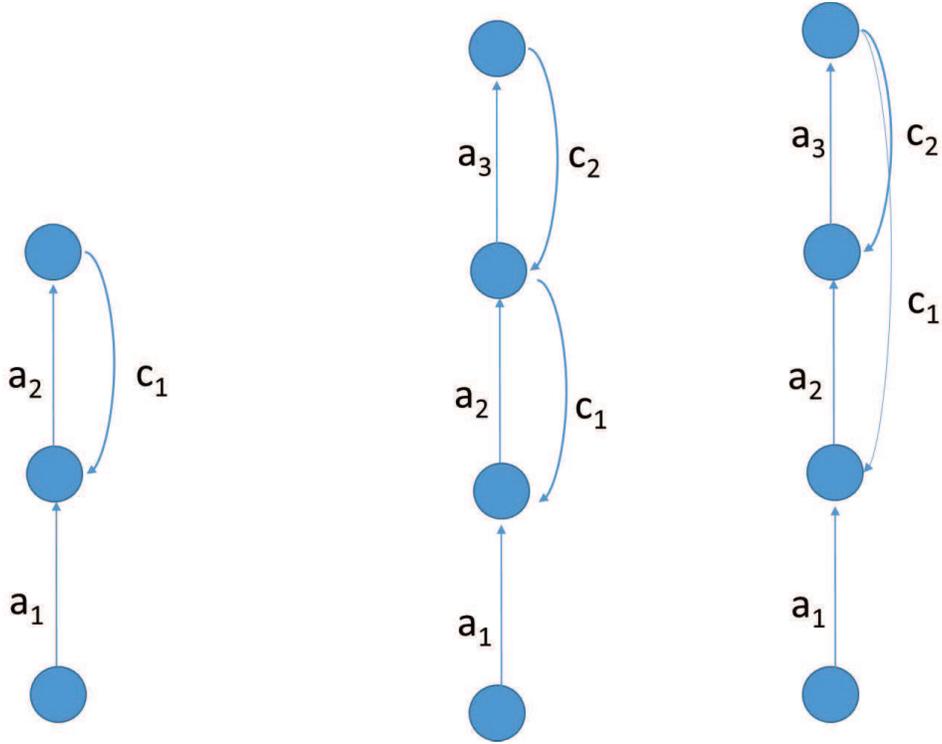}
    \caption{Left: ${\cal A}[1,1,1]$ architecture. The weights $a_1$ and $a_2$ are adjustable, and so is the weight $c_1$ in the learning channel. Right:  ${\cal A}[1,1,1,1]$ architecture in the ARBP and ASRBP cases. The weights $a_1,a_2$, and $a_3$ are adjustable, and so are the weights $c_1$ and $c_2$ in the learning channel. 
    }
    \label{fig:A111}
\end{figure}

{\bf Derivation of the System (ARBP=ASRBP):} The simplest case correspond to a linear ${\cal A}[1,1,1]$ architecture (Figure \ref{fig:A111}). Let us denote by $a_1$ and $a_2$ the weights in the first and second layer, and by $c_1$ the random weight of the learning channel. In this case, we have $O(t)=a_1a_2I(t)$ and the learning equations are given by:

\be
\begin{cases}
\Delta a_1 = \eta c_1(T-O)I=\eta c_1(T-a_1a_2I)I\\
\Delta a_2 = \eta(T-O) a_1I = \eta (T-a_1a_2I) a_1I\\
\Delta c_1= \eta (T-O)a_1I= \eta (T-a_1a_2I) a_1I
\end{cases}
\label{eq:lin1}
\ee
When averaged over the training set:
\be
\begin{cases}
E(\Delta a_1 )= \eta c_1 E(IT) -\eta c_1a_1a_2 E(I^2)=\eta c_1 \alpha -\eta c_1a_1a_2 \beta\\
E(\Delta a_2 )= \eta a_1 E(IT)-\eta a_1^2a_2 E(I^2)=
\eta a_1 \alpha -\eta a_1^2a_2 \beta\\
E(\Delta c_1) = \eta a_1 E(IT)-\eta a_1^2a_2 E(I^2)=
\eta a_1 \alpha -\eta a_1^2a_2 \beta
\end{cases}
\label{eq:lin2}
\ee
where $\alpha=E(IT)$ and $\beta = E(I^2)$.
With the proper scaling of the learning rate ($\eta=\Delta t$) this leads to the non-linear system of coupled differential equations for the temporal evolution of $a_1$, $a_2$, and $c_1$ during learning:

\be
\begin{cases}
\frac{da_1}{dt}= \alpha c_1  - \beta c_1a_1a_2 =c_1(\alpha - \beta a_1a_2)\\
\frac{da_2}{dt} = \alpha a_1- \beta a_1^2a_2 = a_1(\alpha-\beta a_1a_2)\\
\frac{dc_1}{dt}= \alpha a_1- \beta a_1^2a_2 = a_1(\alpha-\beta a_1a_2)
\end{cases}
\label{eq:lin3}
\ee
Note that the dynamic of $P=a_1a_2$ is given by:

\be
\frac{dP}{dt}=a_1\frac{da_2}{dt}+a_2\frac{da_1}{dt}=(a_1^2+a_2c_1)(\alpha-\beta P)
\label{eq:lin4}
\ee
The error is given by:

\be 
{\cal E}=\frac{1}{2} E(T-PI)^2=\frac{1}{2}E(T^2) + \frac{1}{2}P^2\beta -P \alpha=\frac{1}{2}E(T^2) +\frac{1}{2\beta}(\alpha-\beta P)^2-\frac{\alpha^2}{2\beta}
\label{eq:error1}
\ee
and:

\be 
\frac{d {\cal E}}{dP}=-\alpha+\beta P \quad
{\rm with} \quad \frac{\partial {\cal E}}{\partial a_i}=( -\alpha + \beta P )\frac{P}{a_i}
\label{eq:error2}
\ee
the last equality requires $a_i \not = 0$.

\null\par
\noindent
{\bf Theorem 2:}
Starting from any initial conditions the system converges to a fixed point, corresponding to a global minimum of the quadratic error function. All the fixed points are located on the hyperbolas given by $\alpha-\beta P=0$ and are global minima of the error function. For any starting point, the system reduces to a one-dimensional differential equation $da_2/dt=Q(a_2)$ where $Q$ satisfies the conditions of Theorem 1 and its leading term is a third degree
monomial with negative coefficient $-\beta$. As a result $a_2$ converges to a root $r$ of $Q$, $a_1$ converges to $\alpha/(\beta r)$ and $c_1$ converges to $r+c_i(0)-a_2(0)$. Thus if the initial conditions of $c_1$ and $a_2$ are close, they will converge to similar values after learning.

\null\par
\noindent{\bf Proof:}
In this case, the critical points for $a_1$ and $a_2$ are given by:

\be
P=a_1a_2=\frac{\alpha}{\beta} =\frac{E(IT)}{E(I^2)}
\label{eq:lin5}
\ee
which corresponds to two hyperbolas in the two-dimensional $(a_1,a_2)$ plane, in the first and third quadrant for $\alpha=E(IT)>0$.
{\it Note that these critical points do not depend on the feedback weight $c_1$}.
All these critical points correspond to global minima of the error function ${\cal E}={1 \over 2} E[(T-O)^2]$.
Now note that the differential equations for $a_2$ and $c_1$ are identical:

\be
\frac{da_2}{dt}=\frac{dc_1}{dt} \quad
{\rm and \;\; thus} \quad   c_1=a_2+K_1
\label{eq:ad1}
\ee
where $K_1=c_1(0)- a_2(0)$ is a constant depending only on the initial conditions. In addition:

\be
a_1\frac{da_1}{dt}=c_1\frac{da_2}{dt}=(a_2+K_1)\frac{da_2}{dt}
\label{eq:ad22}
\ee
resulting in:

\be
a_1^2=(a_2+K_1)^2+J_1=a_2^2+2K_1a_2+J_1
\label{eq:ad2}
\ee
where $J_1$ depends only on the initial conditions.
Thus, by substituting this value in the differential equation for $a_2$ it is easy to see that it has the form: $da_2/dt=Q(a_2)$ where $Q$ is a functions that satisfies Theorem 1 and its leading term is a monomial of degree 3 in $a_2$ with a negative leading coefficient equal to $-\beta $ (we exclude the trivial case where $\beta=0$ corresponding to a single input equal to 0). Thus $a_2$ is convergent to a fixed point, and so are $a_1$ and $c_1$, and they converge to the values given in the theorem.

\par\null
\noindent
{\bf Derivation of the System (STDP rule):}
This learning rule correspond to the system:

\be
\begin{cases}
\Delta a_1 = \eta Ic_1(T-O)\\
\Delta a_2 = \eta (a_1I+c_1O)(T-O)\\
\Delta c_1= \eta O c_1(T-O)
\end{cases}
\label{eq:}
\ee
As usual, taking expectations, this leads to the system of differential equations:

\be
\begin{cases}
\frac{da_1}{dt} =c_1(\alpha-\beta P) \\
\frac{d a_2}{dt} = (a_1+c_1P)(\alpha-\beta P)\\
\frac{d c_1}{dt}= c_1P(\alpha-\beta P)
\end{cases}
\label{eq:}
\ee
where $\alpha = E(IT)$, $\beta=E(I^2)$, $P=a_1a_2$, and 
$\gamma=E(T^2)$ (not needed in this version).

\par\null
\noindent
{\bf Theorem 2':}
Starting from any initial conditions the system
has long-term existence and converges to a fixed point, corresponding to a global minimum of the quadratic error function. All the fixed points are located on the hyperbolas given by $\alpha-\beta P=0$ and are global minima of the error function.

\par\null
\noindent
{\bf Proof:}
We assume that $[0,T)$ is the maximum time interval of the solution. Of course, we need to prove that $T=+\infty$. For contradiction, assume that $T<+\infty$.

If $c_1(0)=0$, then $c_1=0$ at all times. From the first equation, $a_1$ is a constant. Then the second equation becomes:

\be
\frac{da_2}{dt}=\alpha a_1(0)-\beta a_1^2(0) a_2.
\label{eq:}
\ee
If $a_1(0)=0$, then $a_2$ is a constant as well. If $a_1(0)\neq 0$, then:

\be
a_2(t)=\frac{\alpha}{\beta a_1(0)}(1-e^{-\beta a_1^2(0)t}) +a_2(0)
e^{-\beta a_1^2(0)t}
\label{eq:}	
\ee
which is also convergent.

The equation is invariant under the transformation $a_1\to -a_1, a_2\to -a_2, c_1\to -c_1$. Therefore, in order to prove the theorem, from now on, we only  need to assume that $c_1(0)>0$. 

We observe that when $c_1(0)>0$, then the function $c_1$ is always positive. This is because:

\be
c_1(t)=c_1(0)e^{\int_0^t P(s)(\alpha-\beta P(s))ds }
\label{eq:}
\ee
We also observe that if $\alpha-\beta a_1(0)a_2(0)=0$, then the system has a constant solution, and hence is convergent. So we only need to consider the case where $\alpha-\beta a_1(0)a_2(0)\neq 0$. By the uniqueness of the ODE solutions, $\alpha-\beta a_1a_2$ will not change sign along the solutions.

We first prove that  $a_1$ is bounded. From the first equation and the fact that $\alpha-\beta a_1a_2$ is either positive or negative, we conclude that $a_1$ is always monotonic. Thus if $a_1$ is unbounded, then either $\alpha-\beta a_1a_2>0$ and $a_1\to+\infty$, or $\alpha-\beta a_1a_2<0$ and
$a_1\to-\infty$.

We consider the equation:

\begin{equation}
\frac{d(a_2-c_1)}{dt}=a_1(\alpha-\beta a_1a_2)
\label{eq:25-1}
\end{equation}

If $\alpha-\beta a_1a_2>0$ and $a_1\to+\infty$, or $\alpha-\beta a_1a_2<0$ and $a_1\to-\infty$, then:
 
\be
a_2\leq \Lambda_1
\label{eq:}
\ee
is bounded from above. From Equation \ref{eq:25-1}, we know that $a_2-c_1$ is monotonically increasing for sufficiently large $t$. Therefore the expression $a_2-c_1$ is convergent.  In particular $a_2-c_1$ is bounded. Thus we have:

\be
|a_2|+c_1\leq |\Lambda_1-a_2-\Lambda|+c_1\leq 2\Lambda_1-a_2+c_1
\label{eq:}
\ee
and $|a_2|+c_1$ is also bounded. Integrating Equation \ref{eq:25-1}, we obtain:

\be
\int_0^Ta_1(\alpha-\beta a_1a_2)<+\infty
\label{eq:}
\ee
Since  $|a_1|\to+\infty$, we have:

\be
\int_0^T|\alpha-\beta a_1a_2|<+\infty
\label{eq:}
\ee
Since $c_1$ is bounded, we have:

\be
\int_0^Tc_1|\alpha-\beta a_1a_2|<+\infty
\label{eq:}
\ee
Using the first equation, we have:
\be
|a_1|\leq |a_1(0)|+\int_0^T c_1|\alpha-\beta a_1a_2|<+\infty
\label{eq:}
\ee
and $a_1$is bounded.

Next, we prove the boundedness and convergence of $c_1$ and $a_2$.
We observe that:
 
\be
\frac{d(\alpha a_1-\beta c_1)}{dt}=c_1(\alpha-\beta a_1a_2)^2\geq 0
\label{eq:}
\ee
Since both $a_1$ and $c_1$ are bounded, the expression $\alpha a_1-\beta c_1$ is convergent. Since $a_1$ is monotonic and is bounded, it must be convergent. Thus $c_1$ must be convergent as well. 

Assume that $\lim_{t\to+\infty} c_1\neq 0$. Then since $a_1$ is bounded, $a_1/c_1\leq \Lambda_2$ is also bounded. It follows that:

\be
\int_0^T a_1(\alpha-\beta a_1a_2)\leq \Lambda_2
\int_0^T c_1(\alpha-\beta a_1a_2)<+\infty
\label{eq:}
\ee
From Equation \ref{eq:25-1}, this implies the boundedness and convergence of $a_2-c_1$. By the monotonicity of $a_2-c_1$, we conclude that $a_2$ is also convergent. 

Finally, we assume that $\lim_{t\to+\infty} c_1=0$. We claim that in this case we have  $\lim_{t\to +\infty} a_1>0$ when $\alpha-\beta a_1a_2>0$, and 
$\lim_{t\to +\infty} a_1<0$ when $\alpha-\beta a_1a_2<0$. To prove the claim, we consider the equation for $a_2$:

\be
\frac{da_2}{dt}=a_1(\alpha-\beta a_1a_2) (1+c_1a_2)
\label{eq:}
\ee
If the claim were false, then:

\be
a_1(\alpha-\beta a_1a_2)\leq 0
\label{eq:}
\ee
It follows that if $a_2$ is positive, then $a_2$ is decreasing hence is bounded and convergent. Otherwise for sufficiently large $t$, we must have $a_2\leq 0$. Thus for sufficiently large $t$, $P(\alpha-\beta P)\geq 0$. Therefore $c_1$ is actually monotonically increasing and hence cannot be convergent to $0$. 

Using the claim, we conclude that $a_2$ is bounded from above. Therefore $a_2-c_1$ is convergent, and hence  $a_2$ is also bounded and is convergent. 

Since the bounds for $a_1,a_2,c_1$ are independent of $T$, the system has long-term existence (which means $T=+\infty$) and is convergent. 

\subsection{Adding Depth: the Linear Chain ${\cal A}[1,1,1,1]$.}

{\bf Derivation of the System (ARBP):} In the case of a linear ${\cal A}[1,1,1,1]$ architecture, for notational simplicity, let us denote by $a_1, a_2$ and $a_3$ the forward weights, and by $c_1$ and $c_2$ the random weights of the learning channel (note the index is equal to the target layer). In this case, we have $O(t)=a_1a_2a_3I(t)=PI(t)$. The learning equations are:

\be
\begin{cases}
\Delta a_1 = \eta c_1c_2(T-O)I=\eta c_1c_2(T-a_1a_2a_3I)I\\
\Delta a_2 = \eta c_2(T-O) a_1I = \eta c_2(T-a_1a_2a_3I) a_1I\\
\Delta a_3= \eta (T-O)a_1a_2I=\eta (T-a_1a_2a_3I)a_1a_2I\\
\Delta c_1= \eta c_2(T-O) a_1I= \eta c_2(T-a_1a_2a_3I) a_1I\\
\Delta c_2= \eta (T-O)a_1a_2I= \eta (T-a_1a_2a_3I) a_1a_2 I
\end{cases}
\label{eq:Alinc1}
\ee

As usual, by averaging over the training set, using a small learning rate, and letting $P=a_1a_2a_3$ gives the system of coupled ordinary differential equations:
\be
\begin{cases}
\frac{da_1}{dt}= c_1c_2(\alpha - \beta P)\\
\frac{da_2}{dt} = c_2a_1(\alpha-\beta P)\\
\frac{da_3}{dt}= a_1a_2(\alpha -\beta P)\\
\frac{dc_1}{dt}=       c_2 a_1     (\alpha -\beta P)  \\
\frac{dc_2}{dt}=  a_1a_2  (\alpha -\beta P)
\end{cases}
\label{eq:ASlinc3}
\ee
The dynamic of $P=a_1a_2a_3$ is given by:

\be
\frac{dP}{dt}=a_1a_2\frac{da_3}{dt}+a_2a_3\frac{da_1}{dt}+a_1a_3 \frac{da_2}{dt}=
(a_1^2a_2^2 + c_1a_2a_3+ c_2 a_1^2a_3)(\alpha-\beta P)
\label{eq:ASlinc4}
\ee

\par\null
\noindent
{\bf Theorem 3:}
Starting from any initial conditions the system converges to a fixed point, corresponding to a global minimum of the quadratic error function. All the fixed points are located on the manifold given by $\alpha-\beta P=0$ and are global minima of the error function. Along any trajectory $c_i=a_{i+1}+K_i$
($i=1,2$) where $K_i=c_i(0)-a_{i+1}(0)$ is a constant that depends only on the initial conditions. Thus if $K_i$ is small, $c_i \approx a_{i+1}$ at all times during learning. Along any trajectory, $a_i^2$ is a quadratic function of $a_{i+1}$, ($i=1,2$).The system can be reduced to the system $da_3/dt=Q(a_3)$
where $Q$ satisfies the conditions of Theorem 1 and its leading term is a monomial of degree seven with negative coefficient equal to $-\beta$. Thus $a_3$ converges to one of the roots $r$ of $Q$ and the other variables also converge to fixed values that can be determined from $r$.

\par\null
\noindent
{\bf Proof:}
The system is solved by noting that:

\be
\frac{dc_i}{dt}=\frac{da_{i+1}}{dt} \quad {\rm or} \quad c_i= a_{i+1}+K_i
\label{eq:Alinc5}
\ee
for $i=1,2$.
In addition:

\be
a_i\frac{da_i}{dt}=c_i\frac{da_{i+1}}{dt}=(a_i+K_i) \frac{da_{i+1}}{dt}
\label{eq:Alinc5}
\ee
for $i=1,2$, and thus:

\be
a_i^2=(a_{i+1}+K_i)^2+J_i=a_{i+1}^2+2K_ia_{i+1}+J_i
\label{eq:Alinc5}
\ee
for $i=1,2$. By substituting in the differential equations for $a_3$ we see that it has the form $da_3/dt=Q(a_3)$ where $Q$ satisfies the conditions of Theorem 1 and its leading term is a monomial of degree 7 with negative coefficient $-\beta$. Thus $a_3$ must converge to a root of $Q$, and therefore the other variables also converge to a fixed point. Note that the weights $c_1$ and $c_2$ track the weights $a_2$ and $ a_3$, and if the initial differences are small, the difference between the final values is equally small.

\par\null
\noindent
{\bf Derivation of the System (ASRBP):} 

\be
\begin{cases}
\Delta a_1 = \eta c_1(T-O)I=\eta c_1(T-a_1a_2a_3I)I\\
\Delta a_2 = \eta c_2(T-O) a_1I = \eta c_2 (T-a_1a_2a_3I) a_1I\\
\Delta a_3= \eta (T-O)a_1a_2I=\eta (T-a_1a_2a_3I)a_1a_2I\\
\Delta c_1= \eta (T-O) a_1I= \eta (T-a_1a_2a_3I) a_1I\\
\Delta c_2= \eta (T-O)a_1a_2I= \eta (T-a_1a_2a_3I) a_1a_2 I
\end{cases}
\label{eq:ASlinc1}
\ee
With the usual assumptions, this leads to the  
system of coupled ordinary differential equations:

\be
\begin{cases}
\frac{da_1}{dt}= c_1(\alpha - \beta P)\\
\frac{da_2}{dt} = c_2a_1(\alpha-\beta P)\\
\frac{da_3}{dt}= a_1a_2(\alpha -\beta P)\\
\frac{dc_1}{dt}=        a_1  (\alpha -\beta P)  \\
\frac{dc_2}{dt}=  a_1a_2  (\alpha -\beta P)
\end{cases}
\label{eq:ASlinc3}
\ee
The dynamic of $P=a_1a_2a_3$ is given by:

\be
\frac{dP}{dt}=a_1a_2\frac{da_3}{dt}+a_2a_3\frac{da_1}{dt}+a_1a_3 \frac{da_2}{dt}=
(a_1^2a_2^2 + c_1a_2a_3+ c_2 a_1^2a_3)(\alpha-\beta P)
\label{eq:ASlinc4}
\ee

\par\null
\noindent
{\bf Theorem 3':} Starting from almost any set of initial conditions (except for a set of measure 0)
the system converges to a fixed point, corresponding to a global minimum of the quadratic error function. All the fixed points are located on the manifold given by $\alpha-\beta P=0$ and are global minima of the error function. Along any trajectory $c_i=a_{i+1}+K_i$
($i=1,2$) where $K_i=c_i(0)-a_{i+1}(0)$ is a constant that depends only on the initial conditions. Thus if $K_i$ is small, $c_i \approx a_{i+1}$ at all times during learning. Along any trajectory, $a_i^2$ is a quadratic function of $a_{i+1}$, ($i=1,2$).The system can be reduced to the system $da_3/dt=Q(a_3)$
where $Q$ is a polynomial of degree seven with negative leading coefficient equal to $-\beta$. Thus $a_3$ converges to one of the roots $r$ of $Q$ and the other variables also converge to fixed values that can be determined from $r$.

\par\null
\noindent
{\bf Proof:}
For the exceptions to the convergence, please see
the more general Theorem 4'.
The system is solved by noting first that:

\be
a_i\frac{da_{i}}{dt}=c_i \frac{dc_i}{dt} \quad{\rm or} \quad c_i^2=a_i^2+K_i
\label{eq:AS4chain1}
\ee
for $i=1,2$, where $K_i=c_i^2(0)-a_i^2(0)$ is a constant depending only on the initial conditions. In addition:

\be
\frac{da_{3}}{dt}=c_i \frac{dc_2}{dt} \quad{\rm or} \quad c_2=a_3+J_2
\label{eq:AS4chain2}
\ee
where $J_2=c_2(0)-a_3(0)$ is a constant depending only on the initial conditions. Combining Equations \ref{eq:AS4chain1} and \ref{eq:AS4chain2}
provides a direct quadratic relationship between $a_2$ and $a_3$:

\be
a_2^2=(a_3+J_2)^2-K_2
\label{eq:AS4chain3}
\ee
From the equations of the system (Equation 
\ref{eq:ASlinc3}) we also have:

\be
\frac {d(a_2+c_2)}{dt}=(a_2+c_2)\frac{dc_1}{dt} \quad {\rm and} \quad \frac{d(a_2-c_2)}{dt}=-(a_2-c_2)\frac{dc_1}{dt}
\label{eq:}
\ee
It follows that:

\be
\frac{d((a_2+c_2)e^{-c_1})}{dt}=0 \quad  {\rm and}
\quad \frac{d((a_2-c_2)e^{c_1})}{dt}=0
\label{eq:}
\ee
Thus there are constants $\mu_1,\mu_2$ such that:

\be
a_2+c_2=\mu_1 e^{c_1}\quad {\rm and} \quad a_2-c_2=\mu_2 e^{-c_1}
\label{eq:}
\ee
We assume both $\mu_1,\mu_2\neq 0$. 
Hence:

\be
a_2=\frac12(\mu_1e^{c_1}+\mu_2e^{-c_1}),\quad c_2=\frac12(\mu_1e^{c_1}-\mu_2e^{-c_1})
\label{eq:}
\ee
Now we know that:

\be
c_1^2=a_1^2+K_1.
\label{eq:}
\ee
We have the following two cases.

\null\par
\noindent
{\bf Case 1:  $K_1\geq 0$.} 
In this case, $c_1=\pm \sqrt{a_1^2+K_1}$ is an analytic function of $a_1$. The equation $a_1'=c_1(\alpha-\beta P)$ can be written as:

\be
\frac{da_1}{dt}=c_1\left(\alpha-\beta a_1 (\frac12(\mu_1e^{c_1}+\mu_2e^{-c_1}))(\frac12(\mu_1e^{c_1}-\mu_2e^{-c_1})-J_2)\right)
\label{eq:}
\ee
We let:

\be
c_1(t)=\pm \sqrt{t^2+K_1}
\label{eq:}
\ee
and define the analytic function:

\be
F(t)=c_1(t)\left(\alpha-\beta t (\frac12(\mu_1e^{c_1(t)}+\mu_2e^{-c_1(t)}))(\frac12(\mu_1e^{c_1(t)}-\mu_2e^{-c_1(t)})-J_2)\right)
\label{eq:}
\ee
For $t$ large, the leading order of the above function is:

\be
-\frac 14 \beta t c_1(t)(\mu_1^2 e^{2c_1(t)}-\mu_2^2 e^{-2c_1(t)})
\label{eq:}
\ee
Obviously, if $c_1(t)\to\pm\infty$, we have:

\be
c_1(t)(\mu_1^2 e^{2c_1(t)}-\mu_2^2 e^{-2c_1(t)})\to +\infty
\label{eq:}
\ee
As a result, we have:

\be
F(+\infty)=-\infty,\quad F(-\infty)=+\infty
\label{eq:}
\ee
and by Theorem 1 the system is convergent.

\null\par
\noindent
{\bf Case 2: $K_1<0$.} In this case, $c_1=\pm\sqrt{a_1^2+K_1}$ is not an analytic function (nor a Lipschitz function) anymore. However, $a_1=\pm\sqrt{c_1^2-K_1}$ is an analytic function. 
We use the equation $dc_1/dt=a_1(\alpha-\beta P)$ and wrote $a_1, a_2, a_3$ in terms of $c_1$. The rest of the proof is similar as the case above, using Theorem 1. 

If one of the $\mu_1, \mu_2$ is zero, for instance $\mu_2=0$,  then $c_2=a_2$. As a result, using Theorem 1,  the system is again convergent to a fixed point. 
A more general proof of this theorem is given in Theorem 4'.

\subsection{The General Linear Chain: ${\cal A}[1, \dots,1]$.}

\begin{figure}[h!]
    \centering
    \includegraphics[width=0.60\textwidth]{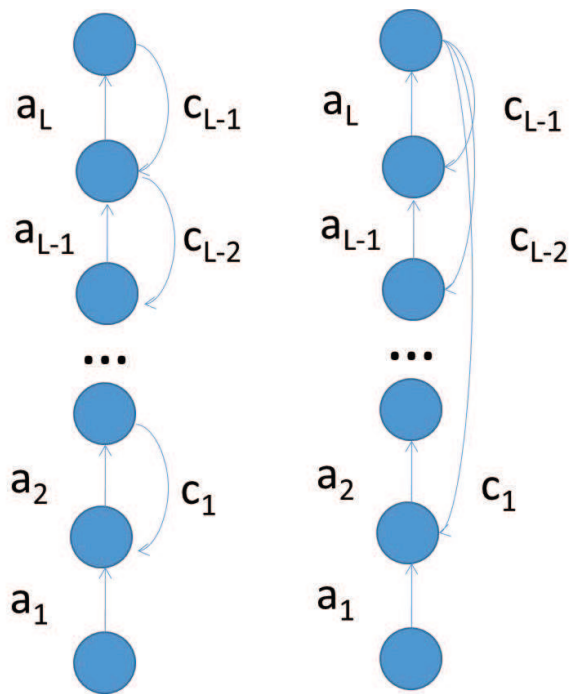}
    \caption{${\cal A}[1,\ldots,1]$ architecture in the ARBP (left) and ASRBP cases (right). The weights $a_i$ are adjustable, and so are the weights $c_i$ in the learning channel. The index of each parameter is associated with the corresponding target layer.   
    }
    \label{fig:A1111}
\end{figure}

\par\null
\noindent
{\bf Derivation of the System (ARBP):}
The analysis can be extended immediately to a linear chain architecture ${\cal A}[1, \dots,1]$ of arbitrary length
(Figure \ref{fig:A1111}). In this case, let $a_1,a_2, \ldots, a_L$ denote the forward weights and $c_1, \dots ,c_{L-1}$ denote the feedback weights. 
Using the same derivation as in the previous cases and letting $O=PI=a_1 a_2 \ldots a_L I$ gives the system: 


\be
\Delta a_i=\eta c_ic_{i+1}\ldots c_{L-1} (T-O) a_1a_2 \ldots a_{i-1}I
\label{eq:Alind0}
\ee
for $i=1,\ldots ,L$.
Taking expectations as usual leads to the set of differential equations:
\be
\begin{cases}
\frac{da_1}{dt}= c_1c_2 \ldots c_{L-1}(\alpha - \beta P)\\
\frac{da_2}{dt} =c_2c_3 \ldots c_{L-1} a_1 (\alpha-\beta P)\\
\ldots \\
\frac{da_{L-1}}{dt}=c_{L-1}a_1a_2 \ldots a_{L-2} (\alpha -\beta P)\\
\frac{da_L}{dt}= a_1\ldots a_{L-1}(\alpha -\beta P)
\end{cases}
\label{eq:Alind3}
\ee
and

\be
\begin{cases}
\frac{d c_1}{dt}= c_2c_3 \ldots c_{L-1}a_1(\alpha - \beta P)\\
\frac{dc_2}{dt} = c_3c_4 \ldots c_{L-1} a_1a_2 (\alpha-\beta P)\\
\ldots \\
\frac{dc_{L-1}}{dt}=a_1a_2 \ldots a_{L-1} (\alpha -\beta P)\\
\end{cases}
\label{eq:Alind31}
\ee
or, in more compact form:

\be
\frac{da_i}{dt}=\prod_{k=i}^{k=L-1} c_k  \prod_{k=1}^{k=i-1} a_k (\alpha -\beta P) \quad {\rm for} \quad i=1, \dots, L\\
\label{eq:Alind41}
\ee
and:

\be
\frac{dc_i}{dt}= \prod_{k=1}^{k=i} a_k \prod_{k=i+1}^{k=L-1} c_k 
(\alpha -\beta P) \quad {\rm for} \quad i=1, \dots, L-1\\
\label{eq:Alind42}
\ee
with $c_L=1$.
As usual, $P=\prod_{i=1}^L a_i$, $\alpha= E(TI)$,
and $\beta=E(I^2)$. 

\par\null
\noindent
{\bf Theorem 4:}
Starting from any initial conditions the system converges to a fixed point, corresponding to a global minimum of the quadratic error function. All the fixed points are located on the manifold given by $\alpha-\beta P=0$ and are global minima of the error function. Along any trajectory $c_i=a_{i+1}+K_i$
($i=1,2,\ldots ,L-1$) where $K_i=c_i(0)-a_{i+1}(0)$ is a constant that depends only on the initial conditions. Thus if $K_i$ is small, $c_i \approx a_{i+1}$ at all times during learning. Along any trajectory, $a_i^2$ is a quadratic function of $a_{i+1}$, ($i=1,2, \ldots,a_{L-1}$).The system can be reduced to the system $da_L/dt=Q(a_L)$
where $Q$ satisfies the conditions of Theorem 1 and its leading terms is a monomial of odd degree $2^{L-1}-1$  with negative leading coefficient equal to $-\beta$. Thus $a_L$ converges to one of the roots $r$ of $Q$ and the other variables also converge to fixed values that can be determined from $r$.

\par\null
\noindent
{\bf Proof:}
As usual, the critical points correspond to the manifold $\alpha-\beta P=0$, and these are all global minima of the error function $\cal E$.
To solve the system, we first have:

\be
\frac{dc_i}{dt}=\frac{da_{i+1}}{dt} \quad {\rm or} \quad c_i=a_{i+1}+K_i
\label{eq:Alongchain1}
\ee
for $i=1,\ldots,L-1$, where $K_i=c_i(0)-a_{i+1}(0)$ is a constant that depends only on the initial conditions. 
This shows that each weight $c_i$ in the learning channel, tracks the corresponding weight $a_{i+1}$ in the forward channel at all times, and the difference is determined only by the initial conditions. If the weights are initialized similarly, for instance by sampling from a Gaussian with mean zero and small standard deviation, then $K_i \approx 0$ and $c_i \approx a_{i+1}$ at all times, including the final state.

Similarly, we have:

\be
a_i\frac{da_i}{dt}=c_i \frac{da_{i+1}}{dt}=(a_{i+1}+K_i)\frac{da_{i+1}}{dt}
\label{eq:Alongchain2}
\ee
for $i=1,\ldots,L-1$, and thus:

\be
a_i^2=(a_{i+1}+K_i)^2+J_i= a_{i+1}^2+2K_ia_{i+1}+J_i
\label{eq:Alongchain3}
\ee
for $i=1,\ldots,L-1$. Substituting in the differential equation for 
$a_L$ gives $da_L/dt=Q(a_L)$ where $Q$ satisfies the conditions of Theorem 1 and its leading term is a monomial of odd degree $2^{L-1}-1$, with negative leading coefficient equal to $-\beta$.

\par\null
\noindent
{\bf Derivation of the System (ASRBP):}
Taking expectations as usual leads to the set of differential equations:
\be
\begin{cases}
\frac{da_1}{dt}= c_1(\alpha - \beta P)\\
\frac{da_2}{dt} =c_2 a_1 (\alpha-\beta P)\\
\ldots \\
\frac{da_{L-1}}{dt}=c_{L-1}a_1a_2 \ldots a_{L-2} (\alpha -\beta P)\\
\frac{da_L}{dt}= a_1\ldots a_{L-1}(\alpha -\beta P)
\end{cases}
\label{eq:ASlind3}
\ee
and:

\be
\begin{cases}
\frac{d c_1}{dt}= a_1 (\alpha - \beta P)\\
\frac{dc_2}{dt} =a_1a_2  (\alpha-\beta P)\\
\ldots \\
\frac{dc_{L-1}}{dt}=a_1a_2 \ldots a_{L-1} (\alpha -\beta P)\\
\end{cases}
\label{eq:ASlind3}
\ee
or, in more compact form:

\be
\frac{da_i}{dt}=c_{i} \prod_{k=1}^{k=i-1} a_k  (\alpha -\beta P) \quad {\rm for} \quad i=1, \dots, L\\
\label{eq:ASlind5}
\ee
and:
\be
\frac{dc_i}{dt}= \prod_{k=1}^{k=i} a_k 
(\alpha -\beta P) \quad {\rm for} \quad i=1, \dots, L-1\\
\label{eq:ASlind6}
\ee
with $c_L=1$.
As usual, $P=\prod_{i=1}^L a_i$, $\alpha= E(TI)$,
and $\beta=E(I^2)$. 

\par\null
\noindent
{\bf Theorem 4':}
 Starting from almost any set of initial conditions (except for a set of measure 0) the system converges to a fixed point, corresponding to a global minimum of the quadratic error function. All the fixed points are located on the manifold given by $\alpha-\beta P=0$ and are global minima of the error function. Along any trajectory $c_i=a_{i+1}+K_i$
($i=1,2,\ldots ,L-1$) where $K_i=c_i(0)-a_{i+1}(0)$ is a constant that depends only on the initial conditions. Thus if $K_i$ is small, $c_i \approx a_{i+1}$ at all times during learning. Along any trajectory, $a_i^2$ is a quadratic function of $a_{i+1}$, ($i=1,2, \ldots,a_{L-1}$).The system can be reduced to the system $da_L/dt=Q(a_L)$
where $Q$ is a polynomial of odd degree $2^{L-1}-1$  with negative leading coefficient equal to $-\beta$. Thus $a_L$ converges to one of the roots $r$ of $Q$ and the other variables also converge to fixed values that can be determined from $r$.

\par\null
\noindent
{\bf Proof:} 
We first note that:

\be 
c_i^2=a_i^2+K_i
\label{eq:}
\ee
for constants $K_i$ that depend only on the initial conditions. First, if some of the $K_i$ are zero, then the result is not true. To see this, 
consider the equation:

\be
f'=f(1+f e^{-2f})
\label{eq:}
\ee
For such an equation, if $f(0)>0$, then $f(t)\to+\infty$ so three is no convergence to a fixed point. 
If we take $c_1=a_1=f$, $a_2=e^{-f}$, and $c_2=a_3=-a_2$, then they satisfy the system of Equation 
\ref{eq:ASlinc3}. This gives a counter example that the system is not convergent. Thus in what follows we prove Theorem 4' under the additional assumption  that all $K_i\neq 0$, which is a minor restriction.

By the Equations \ref{eq:ASlind5} and \ref{eq:ASlind6},
we have:

\be
\frac{d(a_i+c_i)}{dt}=(a_i+c_i) \frac{dc_{i-1}}{dt}\quad {\rm and} \quad \frac{d(a_i-c_i)}{dt}=-(a_i-c_i) \frac{dc_{i-1}}{dt}
\label{eq:}
\ee
for $i\geq 2$. As in the proof of Theorem 3', this implies that there are constants $\mu_i,\nu_i$ such that:

\be
a_i+c_i=\mu_{i-1} e^{c_{i-1}}\quad {\rm and} a_i-c_i=\nu_{i-1} e^{-c_{i-1}}
\label{eq:}
\ee
for $i\geq 2$. Therefore, we have:

\be
a_i=\frac{1}{2}(\mu_{i-1} e^{c_{i-1}}+\nu_{i-1} e^{-c_{i-1}})\quad {\rm and} \quad
c_i=\frac {1}{2}(\mu_{i-1} e^{c_{i-1}}-\nu_{i-1} e^{-c_{i-1}})
\label{eq:}
\ee
for $i\geq 2$.
In particular, this implies:

\be
a_ic_i=\frac 14(\mu_{i-1}^2 e^{2c_{i-1}}-\nu^2_{i-1} e^{-2c_{i-1}})
\label{eq:}
\ee
for $i\geq 2$.
We also observe that $m_i\nu_i=K_i\neq 0$, and thus all $\mu_i,\nu_i\neq 0$. In addition, we also observe the following relations: since $a_1a_1'=c_1c_1'$, there is a constant $K$ such that:

\be
c_1^2=a_1^2+K
\label{eq:}
\ee
Since $a_L'=c_{L-1}'$, we must have:

\be
a_L=c_{L-1}+J
\label{eq:}
\ee
for some constant $J$. By the above relations, we know that: $ a_2,\cdots, a_L$, and their product, are 
analytic functions of $c_1$. We now consider two cases.

\par\null
\noindent
{\bf Case 1: $K\geq 0$.} In this case, $c_1=\pm\sqrt{a_1^2+K}$ is an analytic function of $a_1$. As a result,
the right-hand side of the equation:

\be
a_1'=c_1(\alpha-\beta P)
\label{eq:}
\ee
is an analytic function of $a_1$. We let:

\be
F(a_1)=c_1(\alpha-\beta P)
\label{eq:}
\ee
In order to prove the  long-time existence and the convergence of the system, we just need 
to prove that:

\be
F(+\infty)<0,\quad F(-\infty)>0
\label{eq:}
\ee
because of Theorem 1. 
If all $\mu_i,\nu_i$ are non zero, then 
when $a_1\to\pm \infty$,  all of the $a_i, c_i$ go to infinity. Moreover:

\be
\lim_{c_i\to\pm\infty}\frac{a_ic_i}{c_{i-1}}=+\infty
\label{eq:}
\ee
for $i\geq 2$ by the above formula for $a_ic_i$. It follows that :

\be
a_L\cdots a_2/c_1=
\frac{(c_{L-1}+J)a_{L-1}}{c_{L-2}}\cdot\frac{c_{L-2}a_{L-2}}{c_{L-3}}\cdots\frac{c_2a_2}{c_1}
\to +\infty
\label{eq:}
\ee
as $a_1\to\pm \infty$. 
Obviously, this implies that $F(+\infty)=-\infty$, and $F(-\infty)=+\infty$. 

\par\null
\noindent
{\bf Case 2: $K<0$.} In this case, $a_1=\pm\sqrt{c_1^2-K}$ is an analytic function. We can 
express all the functions $a_i, c_i$ as analytic functions of $c_1$ so that we have: 

\be
c_1'=G(c_1)=a_1(\alpha-\beta P)
\label{eq:}
\ee
By using the same method as above, one can show that, for any initial values, this equation is convergent.

\subsection{Adding Width (Expansive): ${\cal A}[1,N,1]$ }

\begin{figure}[h!]
    \centering
    \includegraphics[width=0.95\textwidth]{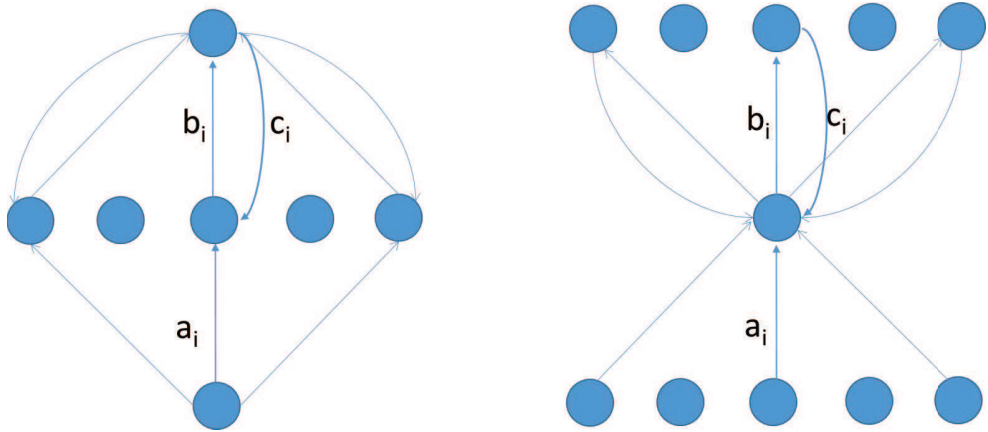}
    \caption{Left: Expansive ${\cal A}[1,N,1]$ Architecture. Right: Compressive ${\cal A}[N,1N]$ Architecture. In both cases, the parameters $a_i$ and $b_i$ are adjustable, and so are the parameters $c_i$ in the learning channel. }
    \label{fig:CompExp}
\end{figure}

{\bf Derivation of the System (ARBP=ASRBP):}
Consider a linear ${\cal A}[1,N,1]$ architecture (Figure \ref{fig:CompExp}). For notational simplicity, we let $a_1,\ldots ,a_N$ be the weights in the lower layer, $b_1,\ldots ,b_N$ be the weights in the upper layer, and $c_1,\ldots ,c_N$ the random weights of the learning channel.
In this case, we have $O(t)=\sum_i a_ib_iI(t)$. We let $P=\sum_i a_ib_i$. The learning equations are:

\be
\begin{cases}
\Delta a_i = \eta c_i(T-O)I=\eta c_i(T-\sum_i a_ib_iI)I\\
\Delta b_i = \eta(T-O) a_iI = \eta (T-\sum_i a_ib_iI) a_iI\\
\Delta c_i=\eta(T-O) a_iI=\eta (T-\sum_i a_ib_iI) a_iI
\end{cases}
\label{eq:Alinb1}
\ee
When averaged over the training set:
\be
\begin{cases}
E(\Delta a_i )= \eta c_i E(IT) -\eta c_i P E(I^2)=\eta c_i \alpha -\eta c_iP \beta\\
E(\Delta b_i )= \eta a_i E(IT)-\eta a_i P  E(I^2)=
\eta a_i \alpha -\eta a_iP \beta\\
E(\Delta c_i)=\eta a_i E(IT)-\eta a_i P  E(I^2)=
\eta a_i \alpha -\eta a_iP \beta
\end{cases}
\label{eq:Alinb2}
\ee
where $\alpha=E(IT)$ and $\beta = E(I^2)$.
With the proper scaling of the learning rate ($\eta=\Delta t$) this leads to the non-linear system of coupled differential equations for the temporal evolution of $a_i,b_i$ and $c_i$ during learning:

\be
\begin{cases}
\frac{da_i}{dt}= \alpha c_i  - \beta c_iP =c_i(\alpha - \beta P)\\
\frac{db_i}{dt} = \alpha a_i- \beta a_iP = a_i(\alpha-\beta P)\\
\frac{dc_i}{dt} = \alpha a_i- \beta a_iP = a_i(\alpha-\beta P)
\end{cases}
\label{eq:linb3}
\ee
The dynamic of $P=\sum_i a_ib_i$ is given by:

\be
\frac{dP}{dt}=\sum_i a_i\frac{db_i}{dt}+b_i\frac{da_i}{dt}=(\alpha-\beta P)\sum_i[b_ic_i+a_i^2]
\label{eq:linb4}
\ee

\par\null
\noindent
{\bf Theorem 5:} Starting from almost any set of initial conditions (except for a set of measure 0) the system converges to a fixed point, corresponding to a global minimum of the quadratic error function. Besides the trivial fixed points associated with $a_i=c_i=0$ for $i=1,\ldots, N$, all the other fixed points are located on the manifold given by $\alpha-\beta P=0$ and are global minima of the error function. Along the trajectories, $c_i$ tracks $b_i$ in the sense that $c_i=b_i+K_i$ where $K_i=c_i(0)-b_i(0)$, and $a_i$ is a quadratic function of $b_i$: $a_i^2=b_i^2+2K_ib_i+J_i$ where $J_i$ is a constant.

\par\null
\noindent
{\bf Proof:}
We first have:

\be
\frac{dc_i}{dt}=\frac{db_i}{dt} \quad {\rm or} \quad c_i=b_i+K_i
\label{eq:Aexpansive1}
\ee
for $i=1,\ldots,N$, where $K_i=c_i(0)-b_i(0)$ is a constant that depends only on the initial conditions. Throughout learning $c_i$ tracks $b_i$ and they end up being essentially equal if $K_i$ is small at initialization.
Furthermore: 

\be
a_i\frac{da_i}{dt}=c_i\frac{db_i}{dt}=(b_i+K_i)\frac{db_i}{dt}
\label{eq:Aexpansive2}
\ee
for $i=1,\ldots,N$. This yields:

\be
a_i^2=b_i^2+2K_ib_i+J_i
\label{eq:Aexpansive3}
\ee
for $i=1,\ldots,N$, where $J_i$ is a constant that depends only on the initial conditions ($J_i=a_i^2(0)-b_i^2(0)-2K_ib_i(0)=a_i^2(0)+b_i^2(0)-2c_i(0)b_i(0)$).
We now let $S_a=\sum a_i^2$ and $S_b=\sum_i b_i^2$ be the square norm of the vectors $a=(a_i)$ and $b=(b_i)$. Likewise, let $U_a=\sum_i K_ib_i$ and $U_b=\sum_i K_ib_i$
be the dot product of the vectors $a$ and $b$ with the constant vector $K=(K_i)$, and let $K=\sum K_i^2$ denote the square norm of the constant vector $(K_i)$.
We then have:

\be
\frac{dP}{dt}=\sum_i a_i \frac{db_i}{dt}+ b_i \frac{da_i}{dt}=
(\alpha-\beta P)(S_a+S_b+U_b)
\label{eq:Aexpansive4}
\ee
and:

\be
\frac{dS_a}{dt}=2\sum_i a_i\frac{da_i}{dt}=2(\alpha-\beta P)\sum_i a_ic_i=
2(\alpha-\beta)(P+U_a)
\label{eq:Aexpansive5}
\ee
and:

\be
\frac{dS_b}{dt}=2\sum_i b_i\frac{db_i}{dt}=2(\alpha-\beta P)\sum_i a_ib_i=
2(\alpha-\beta P)P
\label{eq:Aexpansive6}
\ee
and: 

\be
\frac{dU_a}{dt}=(\alpha-\beta P)\sum_i K_ib_i +K_i^2=(\alpha-\beta P)(U_b+K)
\label{eq:Aexpansive7}
\ee
and finally:

\be
\frac{dU_b}{dt}=(\alpha-\beta P)\sum_i K_ia_i=(\alpha-\beta P)U_a
\label{eq:Aexpansive8}
\ee
It can be shown that this system of differential equations in five variables is convergent. For instance, in the case where the system is initialized symmetrically ($K_i=0$ for every $i$), we have $S_a=S_b+J$ (where $J=\sum J_i$ is a constant) and the system can be reduced to the two dimensional system:

\be
\begin{cases}
\frac{dP}{dt}= (\alpha - \beta P)(2S_b+J) \\
\frac{dS_b}{dt} = 2(\alpha-\beta P) P
\end{cases}
\label{eq:Aexpansive9}
\ee
This yields:

\be
P\frac{dp}{dt}=S_b \frac{dSb}{dt}+PJ
\label{eq:Aexpansive10}
\ee
Assuming $J=0$ too, we have $P^2+H=S_b^2 $ for some constant $H=S_b^2(0-)P^2(0)$. In this case, at all times we also have $a_i^2=b_i^2$ for every $i$, and therefore $S_a=S_b$ at all times. Since $(ab)^2=S_aS_b \cos^2 (a,b)$,
we must have at all times $P^2=S_b^2 \cos^2(a,b)$, and therefore $H \geq 0$ with $H=0$ if and only if $a(0)=b(0)$ or $a(0)=-b(0)$. In this case, we can write a differential equation in $P$ alone:

\be
\frac{dP}{dt}=2(\alpha - \beta P)\sqrt{P^2+H}
\label{eq:Aexpansive11}
\ee
If $H>0$ this equation easily satisfies the conditions of Therem 1 and converges to the only fixed point $P=\alpha/\beta$. As a result $S_b$ and $S_a$ are also convergent, and so are all the $a_i$ and $b_i$. The special case where $H=0$, corresponding to the initial conditions $a_i(0)=b_i(0)$ for every $i$, or $a_i(0)=-b_i(0)$ for every $i$, is easily handled separately.

To deal with the more general case, $\xi=\alpha-\beta P$. Then the system can be written as:

\be
\begin{cases}
A'=\xi C\\
B'=\xi A
\\
C'=\xi A
\end{cases}
\label{eq:}  
\ee
where $A,B,C$ are column vectors. Obviously, $C'=B'$. So there is a constant vector $K$ such that:

\be
C=B+2K.
\label{eq:}  
\ee

Let $R=\frac 12 (A+B)+K, S=\frac 12(A-B)+K$. Then we have:

\be
R'=\xi R \quad {\rm and} \quad S'=-\xi S
\label{eq:}  
\ee
Therefore, we can assume that:

\be
R=f(t)R_0,\quad S=g(t) S_0
\label{eq:}  
\ee
where $R_0, S_0$ are constant vectors depending only on the initial values of $A,B,C$. We have:

\be
f'=\xi f \quad {\rm and} \quad  g'=-\xi g
\label{eq:}  
\ee
Thus $(fg)'=0$ and $fg=f(0)g(0)=1$. We also observe that:

\be
\sum_{i} a_i b_i=|R-K|^2-|S-K|^2=f^2|R_0|^2 -2fR_0K  -g^2|S_0|^2 +2gS_0K
\label{eq:}  
\ee
Thus the equation for $f$ becomes:

\be
f'=\alpha-\beta(f^2|R_0|^2- 2f R_0K -\frac{1}{f^2}|S_0|^2 + 2 \frac{1}{f}S_0K)f 
\label{eq:}
\ee
If $R_0=0$, then the above system is NOT convergent. However, if $R_0\neq 0$, then we note that $f$ is always positive, because of the equation $f'=\xi f$, and obviously, if $f$ is small $f'>0$. Thus $0$ is not an attracting point. By Theorem 1, the system is convergent. 

\null\par
\noindent
{\bf Remark:} If $R_0=0$, then $g(t)\to 0$. Since $fg=1$, $f$ cannot be convergent in this case.

\subsection{Adding Width (Compressive): ${\cal A}[N,1,N]$ }

{\bf Derivation of the System (ARBP=ASRBP):} 
Consider a linear ${\cal A}[N,1,N]$ architecture (Figure \ref{fig:CompExp}). The on-line learning equations are given by:

\be
\begin{cases}
\Delta a_i= \eta \sum_{k=1}^N c_k(T_k-O_k)I_i
\\
\Delta b_i= \eta (T_i-O_i) \sum_{k=1}^N a_kI_k
\\
\Delta c_i= \eta \sum_{k=1}^N a_kI_k (T_i-O_i)
\end{cases}
\label{eq:}
\ee
for $i=1,\ldots,N$. As usual taking expectations, using matrix notation and a small learning rate, leads to the system of differential equations:

\be
\begin{cases}
\frac{dA}{dt}= C(\Sigma_{TI}-BA \Sigma_{II})
\\
\frac{dB}{dt}= (\Sigma_{TI}-BA \Sigma_{II})A^t
\\
\frac{dC}{dt}=A((\Sigma_{TI}-BA \Sigma_{II})^t
\end{cases}
\label{eq:}
\ee
Here $A$ is an $1 \times N$ matrix, $B$ is an $N \times 1$ matrix, and $C$ is an $1 \times N$ matrix, and $M^t$ denotes the transpose of the matrix $M$. $\Sigma_{II}=E(II^t)$ and 
$\Sigma_{TI}=E(TI^t)$ are $N \times N$ matrices associated with the data.

\null\par
\noindent
{\bf Theorem 6:} At all times:

\be
C=B^t+K
\ee
where $K$ is a constant matrix that depends only on the initial conditions. 

\null\par
\noindent
{\bf Proof:} This results immediately from:

\be
\frac{dC}{dt}=(\frac{dB}{dt})^t
\label{eq:}
\ee
A more general version of this result is given in the next section.

\subsection{The General Linear Case: ${\cal A}[N_0,N_1, \ldots, N_L]$ }

\begin{figure}[h!]
    \centering
    \includegraphics[width=0.80\textwidth]{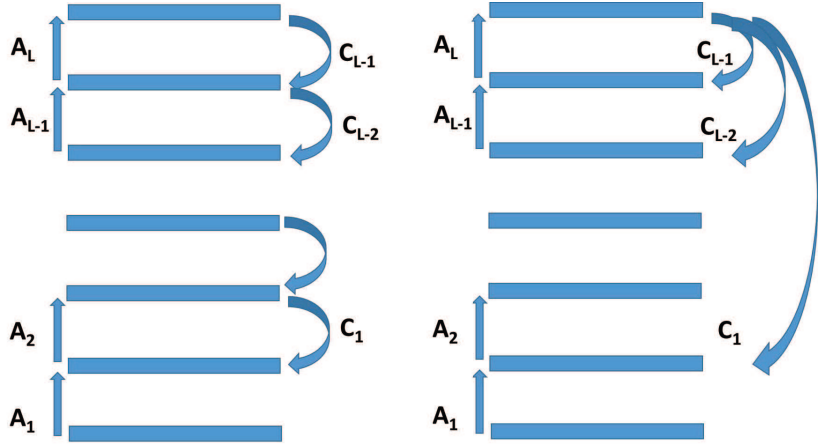}
    \caption{General linear case with an architecture 
    ${\cal A}[N_0, \ldots,N_L]$  for ARBP (left) and ASRBP (right). Each forward matrix $A_i$ is adjustable and of size $N_i \times N_{i-1}$. In ARBP, each feedback matrix $C_i$ is adjustable and of size $N_i \times 
   N_{i+1}$. In ASRBP, each feedback matrix $C_i$ is adjustable and of size $N_i \times N_L$. }
    \label{fig:AGNL}
\end{figure}

{\bf Derivation of the System (ARBP):}
Although we cannot yet provide a solution for this case, it is still useful to derive its equations. We assume a general feedforward linear architecture (Figure \ref {fig:AGNL}) ${\cal A}[N_0,N_1,\ldots, N_L]$ with adjustable forward matrices $A_1, \ldots, A_L$ and adjustable matrices $C_1, \ldots,C_{L-1}$  (and $C_L=Id$) in the learning channel.
Each matrix $A_i$ is of size $N_i \times N_{i-1}$ and, in ARBP, each matrix
$C_i$ is of size $N_i \times N_{i+1}$.
As usual, $O(t)=PI(t)=(\prod_{i=1}^LA_i)I(t)$.

Assuming the same learning rate everywhere, using matrix notation we have:

\be
\Delta A_i =\eta C_iC_{i_1}\ldots C_{L-1} (T-O) (A_{i-1} \ldots A_1 I)^t=\eta C_i\ldots C_{L-1}(T-O)I^tA_1^t \ldots A_{i-1}^t
\label{eq:AGNL0}
\ee
for $i=1, \ldots,L$, and:

\be
\Delta C_i =\eta A_{i-1} \ldots A_1 I [C_{i+1}\ldots C_{L-1}  (T-O)]^t
 =\eta A_{i-1} \ldots A_1 I (T-O)^tC_{L-1}^t \ldots C_{i+1}^t
\label{eq:AGNL00}
\ee
for $i=1,\ldots, L-1$. Taking expectations leads to the system of differential equations:

\be
\frac{d A_i}{dt} = C_i\ldots C_{L-1} (\Sigma_{TI} - P \Sigma_{II}) A_1^t \ldots A_{i-1}^t
\label{eq:AGNL10}
\ee
and

\be
\frac{d C_i}{dt} = A_{i-1}\ldots A_1(\Sigma_{TI} - P \Sigma_{II})^t C_{L-1}^t \ldots C_{i+1}^t
\label{eq:AGNL100}
\ee
with $P=A_L A_{L-1} \ldots A_1$, $\Sigma_{TI}=E(TI^t)$, and $\Sigma_{II}=E(II^t)$. $ \Sigma_{TI}$ is a $ N_L \times N_0$ matrix and 
$\Sigma_{II}$ is a $N_0 \times N_0 $ matrix. In the case of an autoencoder, $T=I$ and therefore $\Sigma_{TI}=\Sigma_{II}$.
Equation \ref{eq:AGNL10} is true also for $i=1$ and $i=L$ with $C_L=Id$ where $Id$ is the identity matrix. From the equations of the system, we immediately have the following Theorem.

\null\par
\noindent
{\bf Theorem 7:} In all cases, $dC_i/dt=dA_{i+1}^t/dt$ for $i+1,\ldots ,L-1$ so that $C_i=A_{i+1}^t+K_i$, where $K_i$ is a constant matrix that depends only on the initial conditions [ $ {(C_i)}_{ij}={(A_{i+1})}_{ji}+{({K_i})}_{ij}$]. If all the entries of $K_i$ are small, which is the case with typical random initialization schemes, the matrix $C_i$ is essentially equal to $A_{i+1}$ at all times during learning, hence the weight are essentially symmetric at all times during learning. Furthermore, if $K_i=0$ for every $i$, then the system correspond to gradient descent on $\cal E$ and thus is convergent.

\par\null
\noindent
{\bf Derivation of the System (ASRBP):}
Note that in the case of ASRBP with backward matrices $C_1,
\ldots, C_{L-1}$, 
each matrix $A_i$ is of size $N_i \times N_{i-1}$ and each matrix
$C_i$ is of size $N_i \times N_L$.
Assuming the same learning rate everywhere, using matrix notation we have:

\be
\Delta A_i =\eta C_i (T-O) (A_{i-1} \ldots A_1 I)^t=\eta C_i(T-O)I^tA_1^t \ldots A_{i-1}^t
\label{eq:ASGLN}
\ee
and

\be
\Delta C_i =\eta (A_{i} \ldots A_1 I)(T-O)^t  =\eta A_{i} \ldots A_1 I(T-O)^t
\label{eq:ASGLN}
\ee
which, after taking averages, leads to the system of differential equations:

\be
\frac{d A_i}{dt} = C_i (\Sigma_{TI} - P \Sigma_{II}) A_1^t \ldots A_{i-1}^t
\label{eq:ASGNL10}
\ee
and:

\be
\frac{d C_i}{dt} =  A_{i-1} \ldots A_1(\Sigma_{TI} - P \Sigma_{II})^t 
\label{eq:ASGNL10}
\ee
with $P=A_L A_{L-1} \ldots A_1$, $\Sigma_{TI}=E(TI^t)$, and $\Sigma_{II}=E(II^t)$. $ \Sigma_{TI}$ is a $ N_L \times N_0$ matrix and 
$\Sigma_{II}$ is a $N_0 \times N_0 $ matrix. In the case of an autoencoder, $T=I$ and therefore $\Sigma_{TI}=\Sigma_{II}$.
Equation \ref{eq:ASGNL10} is true also for $i=1$ and $i=L$ with $C_L=Id$ where $Id$ is the identity matrix. 
Note that we have  $A_i+A_i^T=C_iC_i^T$.

\subsection{ The General Three-Layer Linear Case ${\cal A}[N_0,N_1,N_2]$.}

{\bf Derivation of the System (ASRBP=ARBP):} Here we let $A_1$ be the $N_1 \times N_0$ matrix of weights in the lower layer, $A_2$ be $N_2 \times N_1$ matrix of weights in the upper layer, and $C_1$ the $N_1 \times N_2$ random matrix of weights in the learning channel.
In this case, we have $O(t)=BAI(t)=PI(t))$ and $\Sigma_{II}=E(II^t) $ ($N_0 \times N_0 $) and $\Sigma_{TI}=E(TI^t) $ ($N_2 \times N_1 $).
The system of differential equations is given by:

\be
\begin{cases}
\frac{dA_2}{dt}= (\Sigma_{TI} - P \Sigma_{II})    A_1^t\\
\frac{dA_1}{dt}= C_1 (\Sigma_{TI} - P \Sigma_{II})  \\
  \frac{dC_1}{dt}= A_1(\Sigma_{TI} - P \Sigma_{II})^t   
\end{cases}
\label{eq:ALGN3}
\ee

Note again that we have $dC_1/dt=(dA_2/dt)^t$ and therefore at all time points $C_1=A_2^t+ K$ where $K$ is a constant matrix that depends only on the initial conditions. If $K=0$ at initialization then $C_1=A_2^t$ and the system follows gradient descent on $\cal E$.

\subsection{A Non-Linear Case}

As can be expected, the case of non-linear networks is 
challenging to analyze mathematically. In the linear case, the transfer functions are the identity and thus all the derivatives of the transfer functions are equal to 1 and thus play no role. The simulations reported above provide evidence that in the non-linear case the derivatives of the activation functions play a role in both RBP and SRBP.
Here we study a very simple non-linear case which provides some further evidence.

We consider a simple ${\cal A}[1,1,1]$ architecture, with a single power function non linearity with power $\mu \not = 1$ in the hidden layer, so that $O^1(S)=(S^1)\mu$. The final output neuron is linear $O^2(S^2)=S^2$ and thus the overall input-output relationship is:  $O=a_2(a_1I)^\mu$.
Setting $\mu$ to $1/3$, for instance, provides an S-shaped transfer function
for the hidden layer, and setting $\mu=1$ corresponds to the linear case analyzed in a previous section. The weights are $a_1$ and $a_2$ in the forward network, and $c_1$ in the learning channel. 

\par\null
\noindent
{\bf Derivation of the System With Derivatives (Forward Channel Only):}
When the derivative of the forward activation is included, and the learning channel is adaptive, the system becomes:

\be
\begin{cases}
\frac{da_2}{dt}= a_1^\mu [E(TI^\mu)-a_2a_1^\mu E(I^{2\mu})]=
a_1^\mu(\alpha -\beta a_2a_1^{\mu})
\\
\frac{da_1}{dt}=c_1\mu a_1^{\mu-1}E(TI^\mu)-a_2c_1 \mu    a_1^{2\mu-1}E(I^{2\mu})=a_1^{\mu-1}c_1 \mu
(\alpha-\beta a_2a_1^\mu) \\
\frac{dc_1}{dt}=a_1^\mu[E(TI^\mu)-a_2a_1^\mu E(I^{2\mu})]=a_1^\mu ( \alpha -\beta a_2a_1^\mu)
\end{cases}
\label{eq:wd1}
\ee

\null \par
\noindent
{\bf Theorem 7:} At all times, $c_1=a_2+K_1$ where $K_1=c_1(0)-a_2(0)$ and $c_1$ tracks $a_2$. If $K_1=0$ the system implements gradient descent on $\cal E$ and thus is convergent. If $K_1 \not = 0$ and $\mu$ is an integer equal or greater to 1, then  for any initial conditions the system is convergent. The result remains true for any $\mu >1$ if $K=\mu c_1^2-a_i^2 \geq 0$.

\null \par
\noindent
{\bf Proof:}
The differential equations lead to the coupling:

\be
a_1\frac{da_1}{dt}=(a_2+K_1) \mu \frac{da_2}{dt} \quad
{\rm or} \quad a_1^2=\mu a_2^2+2\mu K_1a_2+J_1
\label{eq:Awd2}
\ee
where $J_1$ is a constant that depends only on the initial conditions. Obviously, for this system, there exist constants $J,K$ such that:

\be
a_2=c_1+J \quad {\rm and} \quad \mu c_1^2-a_1^2=K
\label{eq:}  
\ee
We now assume that $\mu\geq 1$. 

\par\null
\noindent
{\bf Case 1: $\mu$ is a real number.} In this case, to make sure that all the functions $a_1,a_2, c_1$ are positive, we need to assume that $K\geq 0$. We then have:

\be
c_1=\frac{1}{\sqrt\mu}\sqrt{a_1^2+K}
\label{eq:}  
\ee
Then we have:
\be
\frac{da_1}{dt}=\sqrt \mu a_1^{\mu-1}\sqrt{a_1^2+K}\left(\alpha-\beta(\frac{1}{\sqrt\mu}\sqrt{a_1^2+K}+J)a_1^\mu\right)
\label{eq:}  
\ee
By Theorem 1, the system is convergent. 

\null\par
\noindent
{\bf Case 2: $\mu$ is a positive integer.} In this case, if $K\geq0$, the convergence follows from the above case. Now we assume that $K<0$. Then the function:

\be
F(a_1)=\sqrt \mu a_1^{\mu-1}\sqrt{a_1^2+K}\left(\alpha-\beta(\frac{1}{\sqrt\mu}\sqrt{a_1^2+K}+J)a_1^\mu\right)
\label{eq:}  
\ee
is only defined when $a_1^2+K\geq 0$. However, the system has long-term solutions, and 
$a_1^2+K=\mu c_1^2$ must be non-negative. Therefore the system is still convergent. 

\section{Discussion}

In this section, we discuss key issues of biological relevance in the context of some of the vast literature on biological neural systems.

\subsection{ANNs versus BNNs}

A first issue is whether artificial neural networks (ANNs) provide reasonably good approximations of biological neural networks (BNNs). ANNs made of ``McCulloch and Pitts'' neurons and their variants were originally introduced in the 1940s 
\cite{mcculloch:43,Neumann:58}, as simplified models of BNNs capturing the essence of the neuroscience knowledge available at the time. In these ANNs, dendritic integration is modeled linearly by a simple dot product between the vector of stored synaptic weights and the vector of incoming signals and is followed by the application of a typically non-linear function (e.g. threshold, sigmoidal). While ANNs underwent considerable developments in the 1980s, their biological relevance began also to be vigorously criticized, leading some to the extreme position that ANNs have nothing to do with BNNs. In part as a counter-movement against ANNs, this current of thought lead to 
 the development of detailed compartmental models of neurons
\cite{koch1983nonlinear,bower1995book,hines1997neuron,bower2003genesis,carnevale2006neuron}.
Indeed considerable knowledge has accumulated over the years regarding: the complexity  of the geometry of dendritic trees and dendritic integration; the heterogeneity of neurons; the complexity of neural connectivity at multiple scales;  the biophysical complexity of different kinds of ion channels, neurotransmitters, and synapses; the complexity of different kinds of signals carried along the same axons; the role of gene expression in memory and learning, and so forth
(e.g. \cite{felleman1991distributed,markram1998differential,kandel2000principles,guan2002integration,abbott2004synaptic,mayford2012synapses,woodnatureneuroscience12,markov2014anatomy}).

However, even in the 1980s, there were experiments showing that 
ANNs with backpropagation are capable of capturing essential response properties of 
BNNs, for instance in the visual system \cite{zipser1988back}. 
In addition, several articles (e.g.  \cite{olshausen1996emergence}) have shown that in ANNs taylored for vision tasks, learning 
naturally leads to the emergence of arrays of Gabor-like filters that are similar to those found in the corresponding BNNs. These results have been considerably expanded in recent years \cite{yamins2016using}, to the point that there is little doubt that BNNs are capable of capturing at least some essential properties of BNNs in a deep sense,
and the extreme view that ANNs have nothing to do with BNNs is not tenable anymore.

Remarkably, at the most connectionist end of the spectrum, one may actually envision the opposite speculation \cite{baldi2016local}: 
it is BNNs that are trying to approximate ANNs, and not the converse! In this view, pyramidal cells are trying to approximate McCulloch and Pitts neurons and convolutional ANNs, which emerge naturally as mathematical ``canonical'' solutions for problems such as vision. The additional complexity of pyramidal cells and BNNs stems from evolution having to satisfy many other constraints in the physical world (e.g. protein turnover).
Indeed imagine the formidable task of having to build a convolutional neural network for vision using carbon-based computing, or for that matter any other hardware embodiment, with the specifications of being:
1) as accurate as the human system; 2) as fast as the human system (response time scales in the 10-100 ms range); and 3) as durable as the human system (last 100 years). 

In any case, even if most likely the correct answer lies somewhere between the two extremes, ANNs do have something relevant to say about BNNs. If nothing else, ANNs provide the only tractable model we have where information is stored in the connections of a network as opposed to memory addresses, and where storage and processing of information are intimately intertwined rather than being separated, as in digital computers.

\subsection{The Deep Learning Channel in Biology}

A second issue, within the supervised learning framework used in this article, is how the deep learning channel may be implemented in biological systems, and in particular whether BNNs use a Bidirectional, Conjoined, Twin, or Distinct architecture (see Section 3) with possibly various amounts of skipping. 
It must be pointed out first that brains have a large number of feedback connections, as well as intra-layer lateral connections
\cite{felleman1991distributed,o1996biologically,markram1998differential,
kandel2000principles,abbott2004synaptic,markov2014anatomy}.
It is principally for these reasons, together with the relaxation of any symmetry constraints as demonstrated in this article, that in our opinion the Distinct case is the most likely. However this is at best an educated guess and others have argued that more symmetric BNN architectures are possible, or even plausible. Even the most symmetric case corresponding to the Bidirectional architecture cannot be entirely ruled out, as several retrograde signaling mechanisms have been reported in the mammalian brain in the literature. For instance, under certain circumstances, postsynaptic neurons can liberate neurotransmitters that travel in a retrograde direction across the synaptic cleft, where they activate receptors on the presynaptic neuron, causing an alteration in synaptic transmitter release \cite{alger2002retrograde}.
This mechanism seems to play a critical role both in long-term synaptic plasticity as well as in short-term regulation of synaptic transmission. One of the most studied families of retrograde messengers is that of endocannabinoids, which have been shown to mediate the rapid backward suppression of pre-synaptic input, especially in hippocampal pyramidal cells and cerebellar Purkinje cells 
\cite{kreitzer2002retrograde}. Furthermore, in the neocortex we find asymmetrical but also symmetrical synapses \cite{carr1996hippocampal,hendry1983cholecystokinin}--the latter being usually associated with inhibitory mechanisms--with the degree of prevalence of asymmetric or symmetric synapses depending on the specific cortical regions \cite{defelipe1997types}.
Certain authors  (e.g. \cite{o1996biologically}) have also argued that there is some indication that the cortex is roughly symmetrically connected, which would be consistent with the Conjoined architecture discussed above, although these claims have been documented only in some cases and only at macroscopic levels of cortical organization.

\subsection{Supervised Learning and Biological Learning}

A third issue is whether supervised learning captures at least some key features of biological learning or not.
Several authors have discussed the biological plausibility of connectionist learning algorithms \cite{o1998six, thorpe1989biological}
and some have proposed various modifications or interpretations of error backpropagation in order to increase its processing realism (e.g. 
\cite{mazzoni1991more,o1996biologically,NIPS1999_1658,xie2003equivalence}).
It is also relatively easy to extend backpropagation to spiking neurons.
Others have argued that unsupervised forms of deep learning 
may be closer to biological reality 
(e.g. \cite{salakhutdinov2015learning,testolin2016probabilistic}).
Likewise, generative neural network models have been related to neurobiological evidence about sampling-based processing in the cerebral cortex
\cite{fiser2010statistically} and implemented using networks of spiking neurons
\cite{buesing2011neural}. Generative neural networks have been implemented also in neuromorphic chips \cite{pedroni2016mapping,serb2016unsupervised}.
In this context, it should be noted however that many forms of unsupervised computing, such as autoencoders, stacks of autoencoders, variational autoencoders, or adversarial networks require communicating error information, computed in the output layers, back to the deep synapses. Therefore the concept of deep learning channels is readily applicable to these models. 

Finally, reinforcement learning \cite{sutton1998reinforcement} has often been proposed as a more biologically-relevant framework for learning \cite{dayan2002reward,dayan2008reinforcement}. In fact deep learning and reinforcement learning have been combined in applications to produce deep reinforcement learning. This has been done, for instance, for the game of Go. The early work in \cite{baldigoNIPS2006,wu2008learning} used deep learning methods, in the form of recursive grid neural networks, to evaluate the board or decide the next move, and to learn across multiple board sizes. More recently, reinforcement learning combined with massive convolutional neural networks has been used to achieve the AI milestone
of building an automated Go player \cite{silver2016mastering} that can outperform human experts. More generally, in value-based deep reinforcement learning, deep learning is applied to the value function \cite{mnih2015human,bellemare2016increasing,van2016deep,
schaul2015universal,rusu2016policy,blundell2016model}.
In policy-based deep reinforcement learning, deep learning is applied to the policy \cite{silver2014deterministic,lillicrap2016continuous,mnih2016asynchronous, levine2016learning,heinrich2016deep}. Naturally, it is possible to combine both value- and policy-based deep reinforcement learning, together with search algorithms--this is precisly the approach taken in \cite{silver2016mastering} for the game of Go.

In any case, it must be pointed out again that reinforcement learning in a neural machine requires communicating information about rewards to deep synapses, and thus the notion of deep learning channel must apply again in some form. As a minimum, in deep reinforcement learning, the notion of deep learning channels and the results described in this article can be applied directly to the deep neural networks used for the evaluation, or for the policy.

\section{Conclusion}

Learning in the machine is a way of thinking about machine learning that takes into account the constraints that the physical world poses on learning machines, from brains to neuromorphic chips \cite{neftci_event-driven_2017}. Its primary goal is to derive fundamental insights about learning that are largely hardware independent, although ultimately specific analyses must be applied to specific learning systems \cite{baldi2016local}. In some ways, this is similar to the manner in which information theory provides fundamental insights about information and communication, before delving into the details of specific communication channels. In essence, learning in the machine requires putting oneself in the shoes of a physical learning system and its components, such as its neurons and its synapses. Putting oneself in the shoes of a neuron, for instance, reveals why the CONNECTED 
problem\footnote{Given a binary input vector, determine whether the 0's are all adjacent to each other (with or without wrap around). The connectedness makes the problem easy to solve for the human visual system. However, a neuron must learn the particular permutation associated with the ordering of the coordinates.} is much harder than it appears to be to a human observer. Likewise, imagining that neurons have a high rate of failure leads immediately to the dropout learning algorithm
\cite{srivastava_dropout_2014,baldidropout14} where, for each training example, neurons are randomly dropped from the training procedure. Incidentally, note that in standard dropout neurons are assumed to function perfectly at production time, leaving room for additional research. Other examples of ``in the machine'' thinking at the neuronal level include using local connectivity as opposed to full connectivity, or relaxing the exact weight sharing assumption behind convolutional neural networks.

More importantly, putting oneself in the shoes of a synapse provides a better appreciation of the deep learning problem, and leads to the notions of local learning, the stratification of learning rules by their functional complexity, the identification of the fundamental limitations of deep local learning, and to local deep learning and the learning channel \cite{baldi2016local}. In this work, we have further studied the learning channel and its possible embodiments in different architectures:
Bidirectional, Conjoined, Twin, and Distinct, and with or without skip connections. For all these different architectures, we have studied some of the symmetries and similarities between the forward channel and the learning channel. In particular, we have focused on whether non-linear transfer functions and adaptation can occur in the learning channel. Overall, together with random backpropagation and its variants, we have shown through simulations that the learning channel is remarkably robust and can be made to work across almost all combinations of architectures, random weights, 
presence or absence of non-linear transfer functions, and presence or absence of adaptation. The only exception perhaps is when the learning channel is both non-linear and adaptive, in which case learning occurs at the beginning but can then be followed by an unstable regime, suggesting that additional research is needed to better understand this regime, and fine tune the learning rule or other conditions that may reduce its presence.

Finally, we have proven several mathematical results, mostly for linear networks with adaptive learning channels. In general, we find that the learning dynamics converges to an equilibrium and the weights in the learning channel tend to track the weights in the forward channel. More generally, polynomial learning rules in linear networks provide a rich source of polynomial systems of differential equations, and perhaps in time this will help rekindle interest in this important but difficult area of mathematics.

\section*{Acknowledgment}

Work supported in part by NSF grant 
IIS-1550705 and DARPA grant D17AP00002 to PB,
and NSF grant DMS-1510232 to ZL. 
We are also grateful for a hardware donation from NVDIA Corporation. We would like to thank Anton Gorodetski for useful discussions. PB and ZL would like to dedicate this manuscript to the memory of their mothers, as both passed away while it was being written.
The content of this document does not necessarily reflect the position or the policy of the Government, and no official endorsement should be inferred. It is approved for public release; distribution is unlimited.
 
\bibliographystyle{abbrv}
\bibliography{baldi,nn,biblio,AWnetsbib}

\end{document}